\documentclass{article}

    \PassOptionsToPackage{numbers, compress}{natbib}

\usepackage[preprint]{neurips_2024}

\usepackage[utf8]{inputenc} 
\usepackage[T1]{fontenc}    
\usepackage{hyperref}       
\usepackage{url}            
\usepackage{booktabs}       
\usepackage{amsfonts}       
\usepackage{nicefrac}       
\usepackage{microtype}      
\usepackage{xcolor}         
\usepackage[utf8]{inputenc}
\usepackage{wrapfig}

\hypersetup{hidelinks}
\usepackage{tcolorbox}
\usepackage{amsmath}
\usepackage{xspace}
\usepackage{multicol}
\tcbuselibrary{skins} 
\newtcolorbox{dialogbox}{
    enhanced,
    boxrule=1pt, 
    colback=black!10, 
    colframe=black, 
    left=1pt, 
    right=1pt, 
    top=3pt, 
    bottom=3pt, 
}
\newtcolorbox{dialogboxone}{
    enhanced,
    boxrule=1pt, 
    colback=black!10, 
    colframe=black, 
    left=1pt, 
    right=1pt, 
    top=1pt, 
    bottom=1pt, 
    width=0.7\textwidth,
}

\newcommand{\xhdr}[1]{\vspace{0em}\noindent{{\bf #1.}}}
\newcommand{\ie}{\textit{i.e., }}
\newcommand{\eg}{\textit{e.g., }}
\newcommand{\aqua}{\textsc{AQuA}\xspace}
\newcommand{\logiqa}{\textsc{LogiQA}\xspace}
\newcommand{\truthqa}{\textsc{TruthfulQA}\xspace}
\newcommand{\gptthree}{\textsc{Gpt-3.5-Turbo}\xspace}
\newcommand{\gptfour}{\textsc{Gpt-4}\xspace}
\newcommand{\llama}{\textsc{Llama-3-8b-Instruct}\xspace}

\def\genbox#1#2#3#4#5#6{
    \leavevmode\raise#4bp\hbox to#5bp{\vrule height#5bp depth0bp width0bp
    \pdfliteral{q .5 w \csname #2COLOR\endcsname\space RG
                       \csname #3PDF\endcsname{#5}{#6} S Q
             \ifx1#1 q \csname #2COLOR\endcsname\space rg 
                       \csname #3PDF\endcsname{#5}{#6} f Q\fi}\hss}}
\def\trianbox   #1#2{\genbox{#1}{#2}  {trian}    {0}   {5}    {2.5}}

\usepackage{subcaption}

\title{On the Hardness of Faithful Chain-of-Thought Reasoning in Large Language Models}
\author{%
  \textbf{Dan Ley}\thanks{Equal Contribution. Correspondence to Sree Harsha Tanneru <sreeharshatanneru@g.harvard.edu>.} \\
  \texttt{dley@g.harvard.edu} \\
  \And
  \textbf{Sree Harsha Tanneru}\footnotemark[1] \\
  \texttt{sreeharshatanneru@g.harvard.edu} \\  
  \And
  Chirag Agarwal \\
  \texttt{chiragagarwall12@gmail.com} \\
  \And
  Himabindu Lakkaraju \\
  \texttt{hlakkaraju@seas.harvard.edu} \\
  \AND
  \\
  Harvard University\\
  Cambridge, MA 02138 \\
}

\begin{document}

\raggedbottom
\maketitle

\begin{abstract}
    As Large Language Models (LLMs) are increasingly being employed in real-world applications in critical domains such as healthcare, it is important to ensure that the Chain-of-Thought (CoT) reasoning generated by these models faithfully captures their underlying behavior. 
While LLMs are known to generate CoT reasoning that is appealing to humans, prior studies have shown that these explanations do not accurately reflect the actual behavior of the underlying LLMs. In this work, we explore the promise of three broad approaches commonly employed to steer the behavior of LLMs to enhance the faithfulness of the CoT reasoning generated by LLMs: in-context learning, fine-tuning, and activation editing. Specifically, we introduce novel strategies for in-context learning, fine-tuning, and activation editing aimed at improving the faithfulness of the CoT reasoning. We then carry out extensive empirical analyses with multiple benchmark datasets to explore the promise of these strategies. Our analyses indicate that these strategies offer limited success in improving the faithfulness of the CoT reasoning, with only slight performance enhancements in controlled scenarios. Activation editing demonstrated minimal success, while fine-tuning and in-context learning achieved marginal improvements that failed to generalize across diverse reasoning and truthful question-answering benchmarks. In summary, our work underscores the inherent difficulty in eliciting faithful CoT reasoning from LLMs, suggesting that the current array of approaches may not be sufficient to address this complex challenge.
\end{abstract}

\section{Introduction}
\label{sec:intro}
Large Language Models (LLMs) are increasingly being employed in diverse real-world applications ranging from content generation and education to commerce and healthcare~\citep{kaddour2023challenges}. One of the primary reasons behind the widespread adoption of these models is their enhanced reasoning capabilities, which enable them to generate responses that appeal to human end users~\citep{brown2020language,wei2022chain}. Furthermore, these models are also capable of explaining the rationale behind the responses they generate, in a manner that is appealing to humans. 
Despite the aforementioned advantages, LLMs also suffer from some critical drawbacks. For instance, while LLMs are adept at producing explanations that cater to human preferences, recent research~\citep{lanham2023measuring, turpin2023language} demonstrated that the explanations generated by these models -- \eg Chain-of-Thought (CoT) reasoning -- do not \emph{faithfully} capture their underlying behavior. The faithfulness of the generated explanations turns out to be an important desideratum in high-stakes applications such as medical diagnostics and legal counseling. Ensuring the faithfulness of LLM-generated CoT reasoning is crucial for decision-makers, such as doctors, who rely on them to determine if, when, and how much to trust the recommendations made by these LLMs.

Despite the criticality of the faithfulness of LLM-generated reasoning, there is very little research on measuring and enhancing this aspect of LLMs. Recently, \citet{lanham2023measuring} introduced a slew of metrics for measuring the faithfulness of the CoT reasoning generated by LLMs. For instance, they propose an \emph{early answering} metric, which considers a generated CoT to be faithful if truncating that CoT causes the model to change its final response. 
While measuring the faithfulness of an LLM-generated CoT is one critical aspect, another piece of this puzzle is figuring out ways to improve the faithfulness of the CoT reasoning generated by LLMs. While prior works have developed approaches to make CoT more aligned with human understanding or knowledge~\cite{lyu2023faithful}, there are no solutions that focus on improving the faithfulness of LLM-generated CoTs in such a way that they accurately capture the behavior of the underlying model (please refer to Appendix for a more detailed discussion on related work).
Furthermore, it remains unclear how difficult it is to improve the faithfulness of LLM-generated CoT reasoning.
\xhdr{Present work} In this work, we address the aforementioned challenges by exploring the promise of three broad approaches—activation editing, fine-tuning, and in-context learning—to enhance the faithfulness of the CoT reasoning generated by LLMs. Activation editing~\cite{li2024inference} involves probing the internal structures of LLMs and strategically updating them to improve certain properties, while fine-tuning focuses on updating model parameters by leveraging curated datasets. In-context learning, on the other hand, involves providing a handful of samples to the model at inference time to tweak its behavior. These three approaches represent different classes of interventions commonly employed in the literature to steer the behavior of LLMs in a desired direction, such as reducing biases and hallucinations. While these approaches have previously been utilized for various tasks~\cite{tonmoy2024comprehensive,liu2024confronting}, including the reduction of biases and hallucinations, they have not been explored in the context of improving the faithfulness of LLM-generated CoT reasoning.

Here, we introduce novel activation editing, fine-tuning, and in-context learning strategies with the goal of improving the faithfulness of LLM-generated CoT reasoning. Specifically, we introduce an activation editing strategy that involves probing LLMs to first identify a vector/direction that corresponds to faithfulness, and then editing specific attention heads by translating along the identified faithfulness vector. Our fine-tuning and in-context learning strategies involve leveraging the metrics outlined in~\citet{lanham2023measuring} to identify specific instances and their corresponding faithful CoT reasoning, and providing these as inputs to the LLM during the fine-tuning or in-context learning phases, respectively.

Despite the promise of these techniques, our findings reveal that none of them significantly enhance the faithfulness of the CoT reasoning generated by LLMs. While activation editing approach demonstrates limited success in amplifying faithful behavior of CoT reasoning, the fine-tuning and ICL approaches slightly improved CoT faithfulness in controlled scenarios but did not generalize well across diverse datasets. Our results underscore the inherent difficulty in eliciting faithful reasoning from LLMs, suggesting that the current array of techniques available to us is insufficient for addressing this complex challenge. Our research emphasizes the need for fundamentally new methodologies that can delve into the inner workings of LLMs to enhance the faithfulness of LLM-generated CoT reasoning, ensuring that LLMs are not only generating correct responses but also doing so in a manner that faithfully reflects their internal reasoning processes.
\section{Preliminaries}
\label{sec:prelims}
Next, we define the notion of faithfulness we use to quantify the reasoning of LLMs and then discuss some notations used to describe different strategies for eliciting faithful reasoning from LLMs.

\begin{figure}
    \centering
    \includegraphics[width=\textwidth]{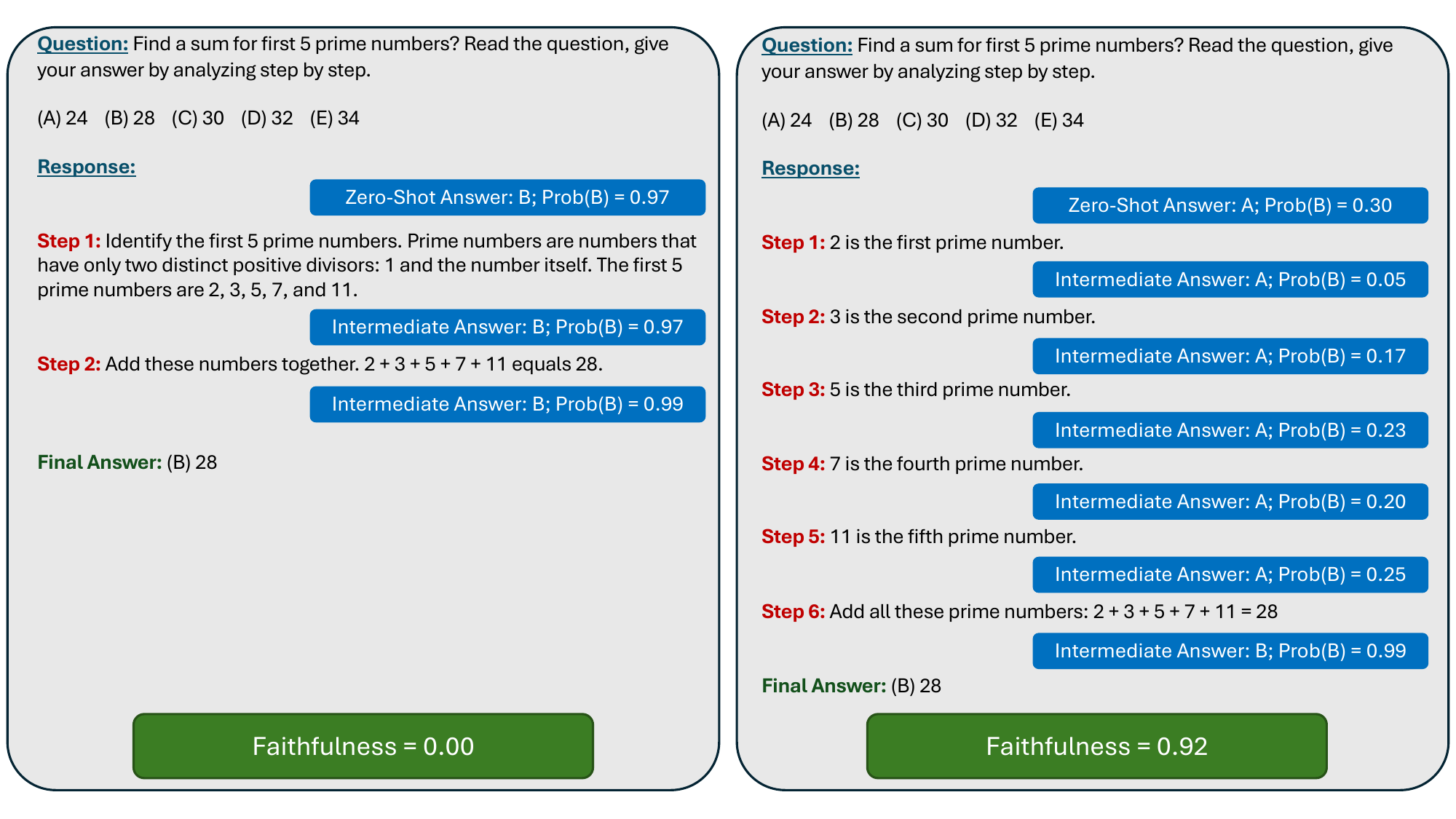}
    \vspace{-0.2in}
    \caption{\footnotesize{Examples for Unfaithful (left) and Faithful (right) explanations generated by state-of-the-art \gptfour (left) and \llama (right) LLMs. The faithfulness score is calculated using the early answering metric proposed in~\citet{lanham2023measuring}. We observe a faithful CoT reasoning gradually improves the prediction probability of the correct answer with an increase in CoT steps.}}
    \label{fig:cot}
\end{figure}

\xhdr{Chain-of-Thought} CoT reasoning in LLMs provides a structured response where the model explicitly generates the step-by-step thought process leading to its final response. This technique is particularly useful in complex reasoning tasks, such as solving math problems or logical question-answering scenarios, and high-stakes decision-making, where transparency in decision-making is crucial. By eliciting intermediate steps, CoT significantly improves the accuracy of LLMs on reasoning tasks and simultaneously leads to greater user trust and understanding. A relevant stakeholder can now see how the LLM processes the input information and relies on it to generate the final output response. See Fig.~\ref{fig:cot} for examples of CoT reasoning.
This CoT reasoning makes the LLM’s process more transparent and easier to trust. Further, this also mimics human problem-solving approaches, allowing for easier debugging and refinement of model reasoning. Formally, let $\mathcal{F}: Q\to A$ denote a large language model that maps a sequence of $n$ input tokens $Q= (q_1, q_2, \dots, q_n)$ to sequence of $m$ answer tokens $A= (a_1, a_2, \dots, a_n)$, where $q_i$ and $a_i$ are text tokens belonging to the model vocabulary $\mathcal{V}$. For CoT reasoning, we append the input tokens $Q$ with a prompt that follows the template: ``\textit{Read the question, give your answer by analyzing step by step, \dots}''.

\xhdr{Notations} For the activation editing of LLMs, we train different linear classifiers $f: x \to y$, where $x\in \mathbb{R}^{d^l_{\text{head}}}$ are the intermediate layer activations of model $\mathcal{F}$ for a given input sequence $X$, $d^l_{\text{head}}$ is the dimension of the model activations at layer $l$ and attention head, and $y$ is the respective label associated with the input. We define sampling functions $S(\tau,~p,~\texttt{mode})$ and $S(\tau,~\texttt{nshot},~\texttt{mode})$ that we use to sample different fine-tuning and in-context examples in our strategies in Sec.~\ref{sec:method}, where $\tau$ determines the temperature parameter of the LLM used to control the randomness in the generated answers by using the probability distribution of each generated token, $p$ denotes the percentage of training examples we use in fine-tuning, $\texttt{nshot}$ denotes the number of training examples we use in the ICL prompting, and $\texttt{mode}$ denotes the sampling technique, \ie whether we want to randomly sample examples from the train split or select the examples with most faithful explanation.

\begin{figure}[h]
    \centering
    \begin{subfigure}{0.5\textwidth}
        \centering
        \begin{dialogboxone}
        \begin{scriptsize}
            \textbf{\textcolor{purple}{\\\textbf{Question (Q):}}} 5! equals what ?\\
            \textbf{\textcolor{purple}{Explanation (E):}}\\
            Step 1: 5! = 1$\times$2$\times$3$\times$4$\times$5.  \\
            Step 2:     1$\times$2$\times$3$\times$4$\times$5 = 120.  \\
            Step 3: Final answer is 120
        \end{scriptsize}
        \end{dialogboxone}
        \vspace{-0.05in}
        \caption{Example of CoT reasoning}
    \label{fig:cotexample}
    \end{subfigure}%
    \begin{subfigure}{0.5\textwidth}
        \centering
        \includegraphics[width=0.65\linewidth]{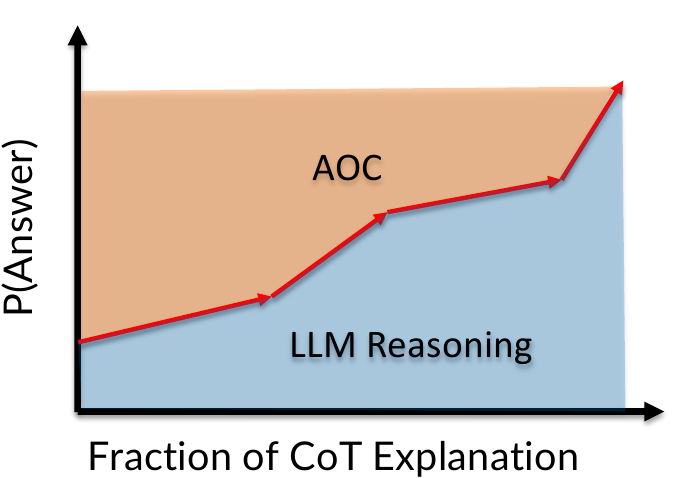}
        \caption{Measuring faithfulness}
        \label{fig:aocplot}
    \end{subfigure}
    \label{fig:introcotexample}
\end{figure}

\xhdr{Measuring Faithfulness} While faithfulness is formally defined as how well an explanation accurately reflects the reasoning process of the underlying LLM, operationalizing this definition in the context of LLMs is non-trivial, partly due to the billion parameter scale, black-box nature of LLMs, and partly due to the internal reasoning (typically a combination of multiple complex nonlinear functions) being in a different representation space from textual CoT reasoning~\citep{agarwal2024faithfulness}. We utilize faithfulness metrics proposed in \citet{lanham2023measuring} that quantify the faithfulness of CoT reasoning from LLMs. Specifically, we employ the \textit{Early Answering} strategy, which evaluates the faithfulness of a CoT by sequentially adding each CoT step to the question and querying the LLM for its answer, conditioned on the truncated set of CoT steps. If the answer from the LLM converges towards the final answer as it encounters more CoT steps, it indicates that the CoT explanation is guiding the answer and is more likely to be faithful.

\looseness=-1 To evaluate the faithfulness of a CoT $E$ shown in Fig.~\ref{fig:cotexample}, the early answering strategy involves providing different truncated versions of $E$ and analyzing how the LLM responds to it. For example, if we provide just the first step of $E$, \ie Prompt = \textit{``5! equals what ? 1: 5! = 1$\times$2$\times$3$\times$4$\times$5.''} and the LLM does not return 120, but it returns 120 when provided with all the steps in $E$, \ie Prompt = \textit{``5! equals what ? Step 1: 5! = 1$\times$2$\times$3$\times$4$\times$5. Step 2: 1$\times$2$\times$3$\times$4$\times$5 = 120. Step 3: So the final answer is 120''}, then we can conclude that $E$ is likely to be faithful. Finally, faithfulness is quantified by the area over the curve (AOC) of explanation fraction vs. the percentage of answers consistent with a full explanation. Note that \citet{lanham2023measuring} measures the faithfulness of CoT reasoning at a dataset level. In contrast, we measure faithfulness of each CoT reasoning using probability scores rather than binary correct or incorrect assessments. Following \cite{lanham2023measuring}, faithfulness is quantified by the area over the curve (AOC) of explanation fraction vs. probability of final answer consistent with a full explanation as shown in Fig.~\ref{fig:aocplot}.

\section{Eliciting Faithful Reasoning from LLMs}
\label{sec:method}

Next, we describe three strategies to improve the faithfulness of CoT reasoning generated by LLMs focusing on different aspects (data, weight, activations) of an LLM, \ie in-context examples (Sec.~\ref{sec:icl}), fine-tuning weights (Sec.~\ref{sec:finetune}), and activation editing (Sec.~\ref{sec:probing}). 

\subsection{Faithful Reasoning via In-Context Learning}
\label{sec:icl}
In contrast to traditional ML approaches that require explicit training or fine-tuning on task-specific data, In-Context Learning (ICL) allows an LLM to generalize and adapt its knowledge by learning patterns from a limited set of demonstrations added within the prompt during inference. ICL is a computationally efficient technique that shows an LLM's capability to transfer knowledge to novel tasks without additional parameter updates and can be used for both open- and closed-source LLMs.

In order to improve the faithfulness of CoT reasoning using ICL, we include demonstrations of faithful CoT in-context before the question. The intuition is that each `faithful' explanation constitutes a set of logical reasoning blocks expressed in natural language, and steering LLMs towards using these filtered faithful reasoning to construct CoT to arrive at an answer, in turn makes their reasoning more faithful. In particular, we consider \(N\) in-context examples, each represented as a triple \((Q_i,  E_i, A_i)\) for \(1 \leq i \leq N\), where \(Q_i\) and \(A_i\) represents the question and answer associated with the \(i\)-th example, while \(E_i\) denotes a `faithful' CoT reasoning for the question \(Q_i\) and answer \(A_i\). Mathematically, we can express the set of \(N\) in-context examples as $\{(Q_1, E_1, A_1), (Q_2, E_2, A_2), \ldots, (Q_N, E_N, A_N)\}$. 

For a given question $Q$, a language model $\mathcal{F}$ and system prompt $S$ to generate CoT reasoning $A_e$ along with an answer $A$, the model $\mathcal{F}$ operates as follows, $\mathcal{F}: (Q + S) \rightarrow (A_e + A)$, whereas in-context learning involves passing in the examples as:
$$\mathcal{F}((Q_1, A_1, E_1) + (Q_2, A_2, E_2), \ldots, + (Q_N, A_N, E_N) + Q + S) = A_e + A,$$
where $N$ demonstrations chosen for ICL impact both the accuracy and faithfulness of answers and CoT reasoning. In order to systematically assess the influence of the specific ICL examples chosen, we propose the following sampling strategies.

\begin{enumerate}
\item[1)] \textbf{Deterministic Uniform (DU).}~Here, we query the LLM deterministically with temperature $\tau = 0$ to yield $(Q, E, A)$ triplets over the full training set. We then uniformly sample $N$ demonstrations for ICL. Mathematically, this can be expressed as $S(\tau{=}0,\  \texttt{nshot}{=}N,\ \texttt{mode}{=}\text{`uniform'})$ (see Sec.~\ref{sec:prelims}).

\item[2)] \textbf{Deterministic Faithful (DF).}~As above, except we select the $N$ most faithful CoT reasoning across the $(Q, E, A)$ triplets, expressed as $S(\tau{=}0,\ \texttt{nshot}{=}N, \ \texttt{mode}{=}\text{`faithful'})$.

\item[3)] \textbf{Stochastic Uniform (SU).}~With this approach, we introduce diversity in eliciting CoT reasoning by sampling at $\tau > 0$, generating 10 samples per question and retaining only the most faithful sample. We then uniformly sample $N$ demonstrations for ICL, expressed as $S(\tau>0, \ \texttt{nshot}{=}N,\ \texttt{mode}{=}\text{`uniform'})$.

\item[4)] \textbf{Stochastic Faithful (SF).}~Here, we combine stochastic sampling with most faithful selection and select the $N$ most faithful demonstrations for ICL, expressed as $S(\tau>0,\ \texttt{nshot}{=}N,\ \texttt{mode}{=}\text{`faithful'})$.
\end{enumerate}

Note that we use these strategies in our empirical analysis and use a superscript $^c$ notation to indicate that only $(Q, E, A)$ triplets with correct answers are used, \eg SF$^c$ indicates that we stochastically generate CoT reasoning, and select the $N$ most faithful triplets that yielded correct answers.

\subsection{Faithful Reasoning via Fine-Tuning}
\label{sec:finetune}
\looseness=-1 Recent progress in LLMs has led to a paradigm shift from the traditional development of models from scratch to an adoption of shared pre-trained LLMs, \eg BERT~\citep{devlin2019bert}, GPT~\citep{brown2020gpt}, Llama~\citep{llama3}, that can readily be fine-tuned for specific downstream applications. 
We utilize a combination of recent techniques like Parameter-Efficient Fine-Tuning (PEFT)~\citep{peft} and Low-Rank Adaptation (LoRA)~\citep{hu2021lora} that allows efficient fine-tuning LLMs on smaller datasets and reduces the number of trainable parameters by learning low-rank adaptation matrices, making the fine-tuning process more memory and computationally efficient while retaining information that is important for downstream performance.

Our exploration of faithful CoT reasoning via fine-tuning is motivated by~\citet{liu2022peftvsicl,ding2023peft} which argues that few-shot PEFT are more effective and cost-efficient as compared to ICL. Hence, we investigate the possible benefits of fine-tuning techniques to elicit more faithful CoT reasoning from LLMs. Our study explores a series of selection strategies aimed at enhancing the faithfulness of CoT reasoning. To this end, we curate a variety of datasets for fine-tuning state-of-the-art LLMs with the goal of fine-tuning LLMs with different question, answer, and CoT reasoning examples and understanding their effects on the faithfulness of CoT reasoning generated by the LLM for test samples during inference. In particular, the strategies we employ for the selection of $(Q, E, A)$ triplets used in finetuning are directly analogous to their ICL counterparts described in Sec.~\ref{sec:icl}:

\begin{enumerate}
\item[1)] \textbf{Deterministic Uniform (DU).}~Selecting all examples (instead of $N$ random examples) for the finetuning dataset: $S(\tau{=}0,\  \texttt{p}{=}100\%,\ \texttt{mode}{=}\text{`uniform'})$.

\item[2)] \textbf{Deterministic Faithful (DF).}~Selecting a percentage of the most faithful examples (instead of the top $N$) for finetuning: $S(\tau{=}0,\ \texttt{p}<100\%, \ \texttt{mode}{=}\text{`faithful'})$.

\item[3)] \textbf{Stochastic Uniform (SU).}~Selecting all examples (instead of $N$ random examples) for the finetuning dataset: $S(\tau>0,\ \texttt{p}{=}100\%, \ \texttt{mode}{=}\text{`uniform'})$.

\item[4)] \textbf{Stochastic Faithful (SF).}~Selecting a percentage of the most faithful examples (instead of the top $N$) for finetuning: $S(\tau>0,\ \texttt{p}<100\%, \ \texttt{mode}{=}\text{`faithful'})$.
\end{enumerate}

As in Sec.~\ref{sec:icl}, the superscript $^c$ notation in the empirical analysis indicates that only $(Q, E, A)$ triplets with correct answers were used for fine-tuning.

\subsection{Faithful Reasoning via Activation Editing}
\label{sec:probing}
Seminal works in explainable artificial intelligence have shown that probing analysis~\citep{alain2016understanding} can find vectors in the activation space of deep neural networks that correlate to specific properties learned by the underlying model. 
Formally, editing activations to steer a LLM's behavior involve two key steps - a probing analysis step to identify which components of the model to intervene on, and an editing step which manipulates the activations at run-time. These two steps are detailed below.

\looseness=-1\xhdr{Step 1: Probing for Faithfulness} Analyzing a model's internal structures, such as individual neurons or specific mechanisms like convolution or attention, can offer insights into the inner workings of LLMs~\citep{li2024inference}. A standard tool to understand a model's inner workings is a ``\textit{probe}''~\citep{alain2016understanding}. Probes are linear classifiers trained on a model's intermediate activations to predict a property like factual correctness, harmful biases, etc. By assessing how well these probes perform, we can infer the extent to which certain types of (mis)information is encoded at different layers or components of the model.

Specifically, we aim to identify attention heads that encode information for faithful reasoning. Using a probing dataset of questions, we collect intermediate activations at all layers and attention heads in a LLM, and create a dataset  $\{(x_i, y_i)\}_{i=1}^{n}$ for each head $h$ and each layer $l$, where $x_i \in \mathbb{R}^{d_{\text{head}}^{l}}$ represents the intermediate activation at a particular layer and attention head of $i^{th}$ question in the probing dataset and $y_i$ represents the faithfulness (measured using approaches described in Sec.~\ref{sec:prelims}) of reasoning generated for $i^{th}$ question. The probing dataset is split into 4:1 training and validation sets, and the probe is a logistic regression classifier $\sigma({\theta_{h}^{l}}^T \boldsymbol{x})$ to predict faithfulness. As faithfulness is a continuous value, we binarize it using median value as threshold. For a model with $L$ layers and $H$ attention heads, we train a total of $L \times H$ linear probes.

\begin{figure}[t]
    \begin{flushleft}
        \footnotesize
        \hspace{1.6cm}\textsc{\aqua}\hspace{3.7cm}\textsc{\logiqa}\hspace{3.2cm}\textsc{\truthqa}
        \vspace{-0.08in}
    \end{flushleft}
    \centering
    \begin{subfigure}[b]{0.32\textwidth}
        \centering
        \includegraphics[width=\linewidth]{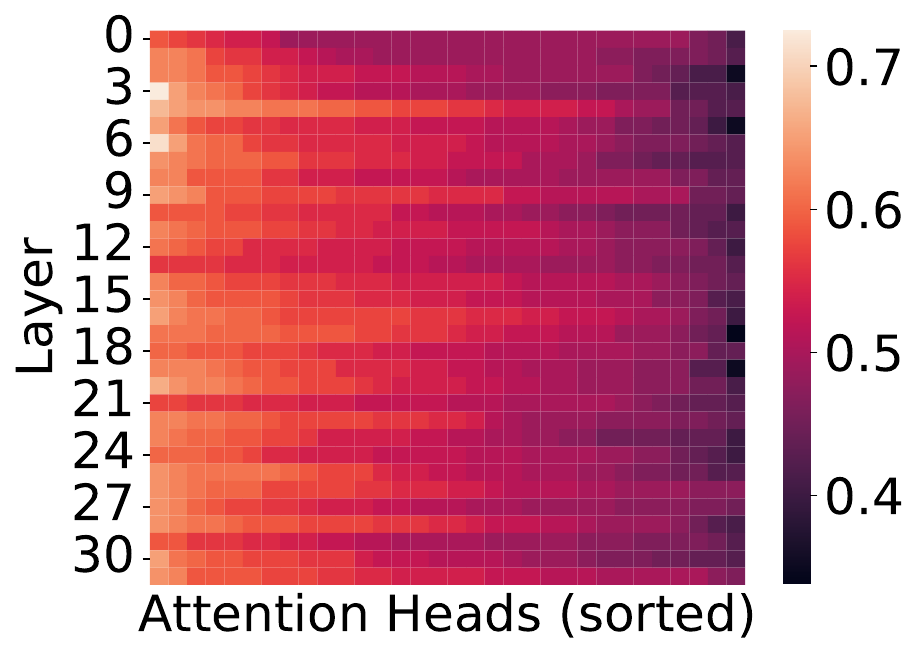}
        \label{fig:subfig1}
    \end{subfigure}
    \hfill
    \begin{subfigure}[b]{0.32\textwidth}
        \centering
        \includegraphics[width=\linewidth]{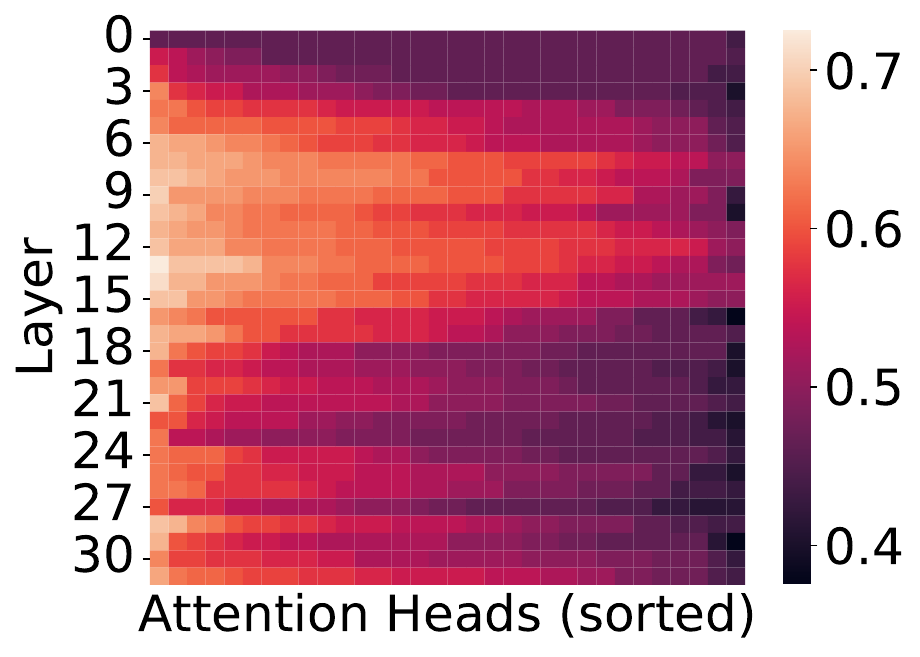}
        \label{fig:subfig2}
    \end{subfigure}
    \hfill
    \begin{subfigure}[b]{0.32\textwidth}
        \centering
        \includegraphics[width=\linewidth]{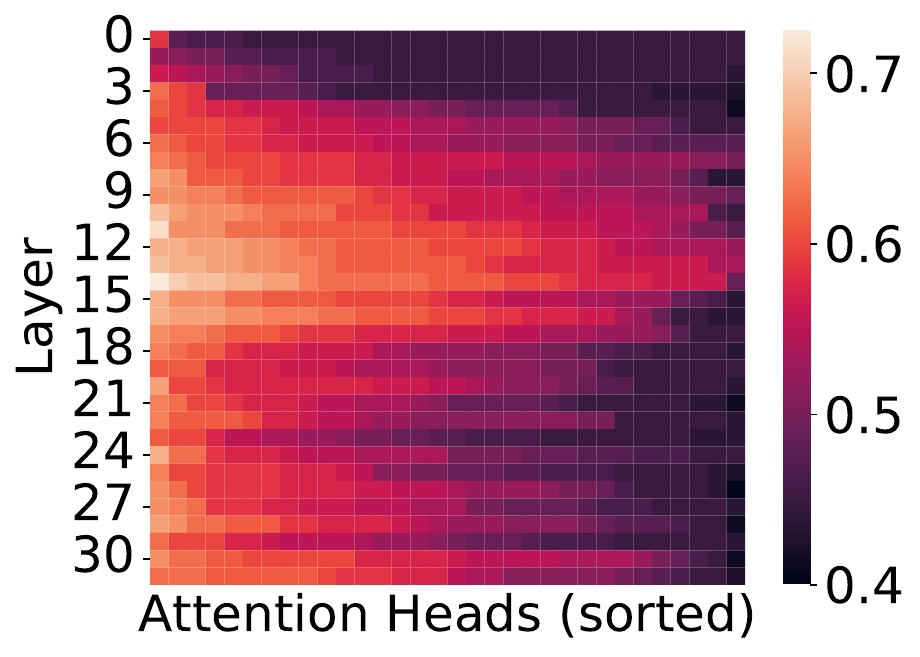}
        \label{fig:subfig3}
    \end{subfigure}
    \vspace{-0.15in}
    \caption{\footnotesize{\looseness=-1 We probe the attention heads across all layers of \llama to assess their predictive power regarding faithfulness. We show the attention heads in each layer sorted by accuracy, clearly indicating that certain attention heads are more responsible for generating faithful explanations.}
    }
    \vspace{-0.1in}
    \label{fig:attentionheads}
\end{figure}

Fig.~\ref{fig:attentionheads} shows the accuracies of linear probes trained on intermediate activations of \llama on three reasoning and math word problem datasets (discussed at detail in Sec.~\ref{sec:expt}). We observe a significant variance in probing accuracy, suggesting that certain attention heads capture more information about faithful reasoning than others.

\xhdr{Step 2: Activation Editing} Activation editing is a technique to control the post-training behavior of models by using steering intermediate activation vectors, \ie simple manipulations like translation, scaling, zeroing out, and clamping, on the internal activations of a model at inference time to achieve a desired outcome. By manipulating specific activations associated with certain behaviors, we can alter the LLM's responses without requiring further training. In our exploration, we apply activation editing to improve the faithfulness of CoT reasoning in LLMs. As shown in \ref{fig:attentionheads}, we first identified specific attention heads that encode more information about faithful CoT reasoning. We then use this information to steer the LLM in the direction that amplifies faithful reasoning. Following~\citet{li2024inference}, we translate the activations of a head by a fixed vector during inference.

To avoid causing OoD inputs for subsequent layers by intervening on every head, we do not translate the activations of all attention heads and focus on the top-K heads ranked by the faithfulness metric (Sec.~\ref{sec:prelims}), thereby intervening on the LLM's behavior in a minimally invasive manner. The parameters of the linear probe classifier indicate the direction in which faithful and unfaithful reasoning are maximally separable. Thus, we translate in the direction represented by the linear probe parameters $\theta$.
\begin{figure}[h]
\vspace{-0.13in}
\begin{equation}    
    \text{Attention}(\mathbf{Q'}, \mathbf{K'}, \mathbf{V'}) = \text{softmax}\left(\frac{\mathbf{Q'} \mathbf{K'}^\top}{\sqrt{d_k}}\right) \mathbf{V'} + \alpha \; \theta_{h}^{l} \; \sigma_{h}^{l},
\end{equation}
\vspace{-0.1in}
\caption{\footnotesize{Attention mechanism used for intervention on attention heads. $\mathbf{Q'}$, $\mathbf{K'}$, and $\mathbf{V'}$ represent query, key, and value matrices respectively. $\alpha$ denotes the intervention strength, $\theta_{h}^{l}$ represents the learned parameters from linear probe at layer $l$ and attention head $h$. $\sigma_{h}^{l}$ is a scaling factor.}}
\label{eq:interventionequation}
\vspace{-0.1in}
\end{figure}
where $\theta_h^l$ denotes the linear probe classifier trained on the activations on layer $l$ and attention head $h$ and $\alpha$ is a hyper-parameter to control the strength of intervention. The direction vector $\theta_h^l$ is scaled by $\sigma_{h}^{l}$, representing the standard deviation of projections of activations in the direction of $\theta_h^l$, ensuring that translation is in the same scale as activations.

\section{Experiments}
\label{sec:expt}
We describe the experimental setup used in our analysis before proceeding to discuss the results.

\subsection{Experimental Setup}
\label{sec:setup}

\xhdr{Datasets} We conduct experiments using math word problems, commonsense reasoning, and factuality-based benchmark datasets. i) the \aqua~\citep{ling_aqua_2017} dataset contains 100,000 algebraic word problems with natural language rationales, where each problem consists of a \textit{question} -- a definition of the problem to solve, \textit{options} -- five possible answer options, where one is correct, \textit{rationale} -- a description of the solution to the problem and \textit{correct} -- a correct option), ii) the \logiqa~\citep{liu_logiqa_2023} consists of 8,678 question-answer instances, covering multiple types of deductive reasoning, where each question has four possible answer options, and iii) the \truthqa~\citep{lin_truthfulqa_2022} dataset contains 817 questions in total, spanning 38 categories (\eg logical falsehoods,
conspiracies, and common points of confusion). Each question comes with an average of $3.2$ truthful
answers, $4.1$ false answers, and a gold standard answer supported by an online source. 



\xhdr{Models} We generate and evaluate the faithfulness of reasoning generated by three large language models -- \llama, \gptthree, and \gptfour.

\looseness=-1\xhdr{Baselines} We use three baselines to evaluate the effectiveness of the ICL, fine-tuning, and activation editing strategies. \textit{1) Zero-shot (ZS):} Here, we assess the accuracy performance of the LLM by just asking the question with invoking CoT reasoning,~\textit{2) Zero-shot CoT (ZS-CoT):} We invoke the CoT reasoning capability in LLMs by prompting the LLM to think step-by-step (see Fig.~\ref{fig:cot}) before answering the question, and ~\textit{3) Ground Truth Answers (GTA):} We provide a random set of ground truth question and answer pairs during ICL and fine-tuning, and evaluate whether it aids the LLM in generating more faithful CoT reasoning.

\subsection{Results}
\label{sec:results}
Next, we discuss the impact of in-context learning, fine-tuning, and activation editing on the faithfulness of CoT reasoning. Our findings indicate that current techniques do not conclusively improve the faithfulness of CoT reasoning in LLMs.

\begin{figure}[t]
    \begin{flushleft}
        \footnotesize
        \hspace{1.9cm}\textsc{\aqua}\hspace{3.9cm}\textsc{\logiqa}\hspace{3.0cm}\textsc{\truthqa}
    \end{flushleft}
    \centering
    \begin{subfigure}[b]{0.32\textwidth}
        \centering
        \includegraphics[width=\linewidth]{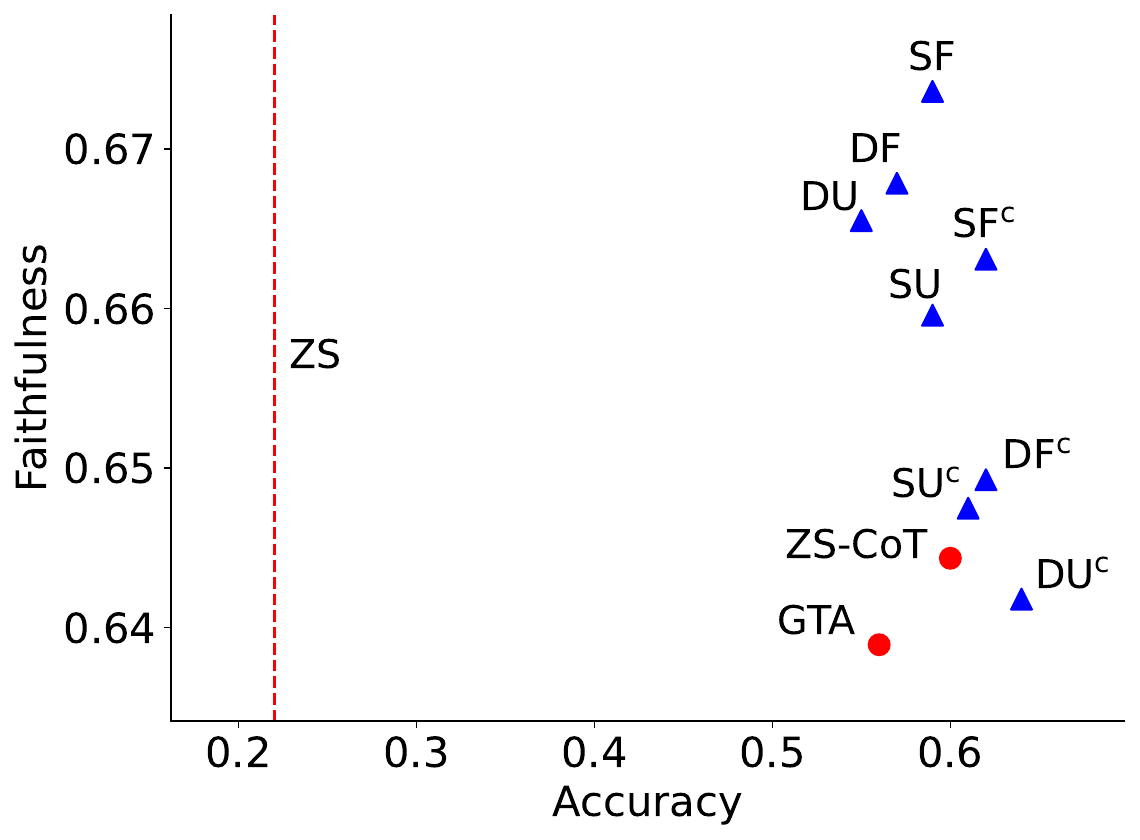}
    \end{subfigure}
    \hfill
    \begin{subfigure}[b]{0.32\textwidth}
        \centering
        \includegraphics[width=\linewidth]{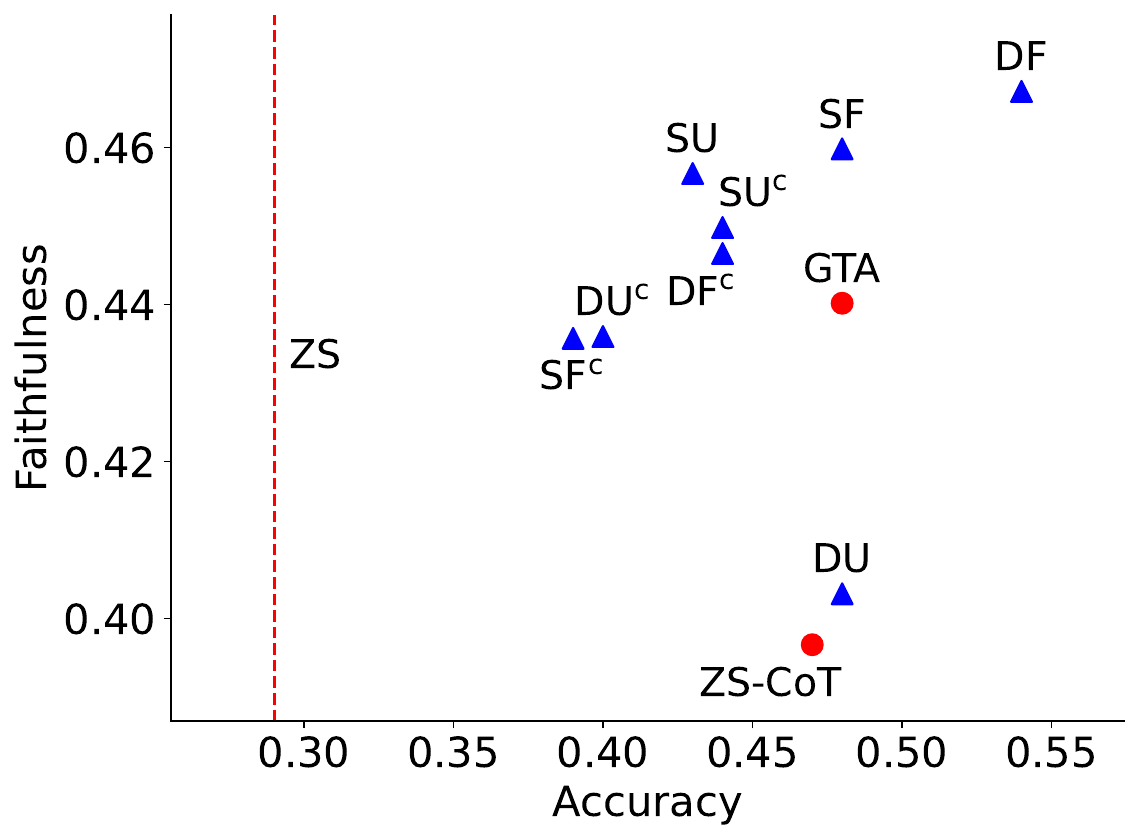}
    \end{subfigure}
    \hfill
    \begin{subfigure}[b]{0.32\textwidth}
        \centering
        \includegraphics[width=\linewidth]{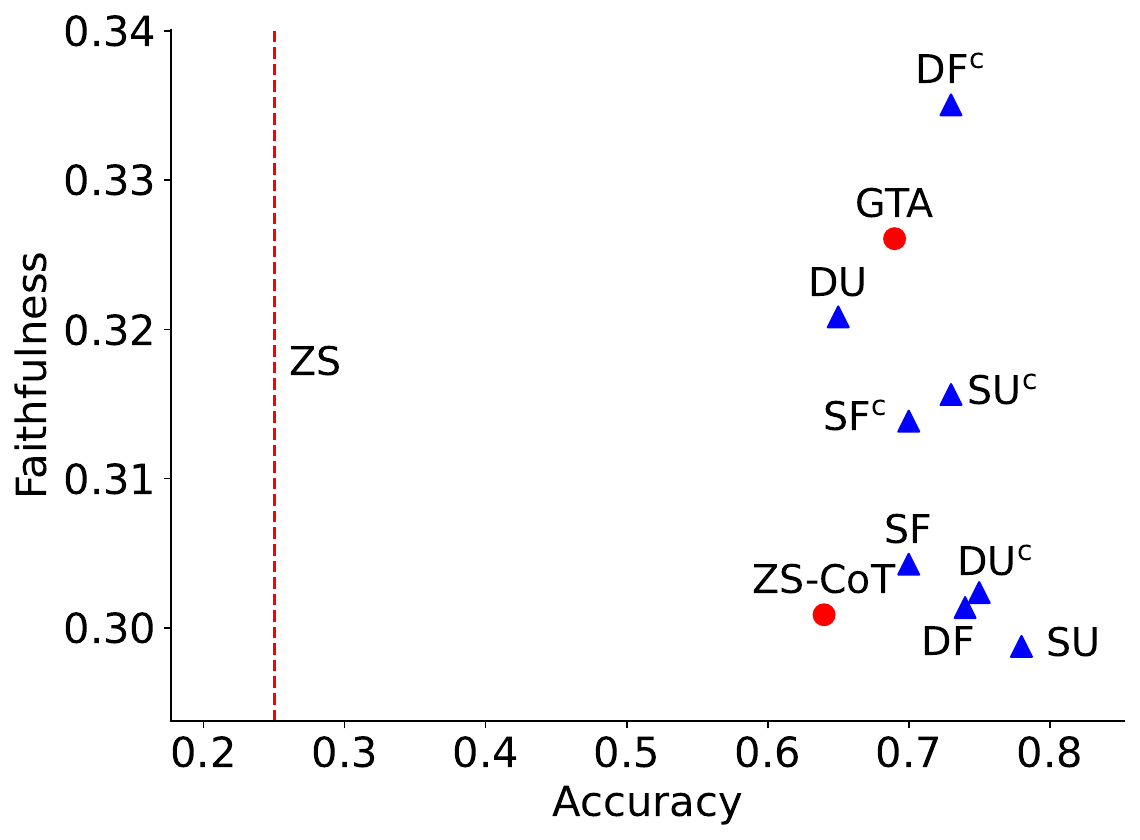}
    \end{subfigure}
    \vspace{-0.1in}
    \caption{\footnotesize{\looseness=-1 Faithfulness vs Accuracy relationship of CoT reasoning generated by \gptthree using different baseline (in red) and \textbf{ICL} strategies (in blue). On average, across all three datasets, we find that deterministic faithful (DF) sampling strategy achieve better accuracy-faithful trade-off.}
    }
    \vspace{-0.1in}
    \label{fig:icl_gpt_faith_acc}
\end{figure}

\subsubsection{In-context Learning Analysis} Using ICL, we aim to address the question: \textit{Can an LLM learn to elicit faithful CoT reasoning by simply looking at some faithful CoT examples during inference?} We investigate this question using the sampling strategies detailed in Sec.~\ref{sec:icl}, and different datasets and LLMs described in Sec.~\ref{sec:setup}.

\begin{figure}[t]
    \begin{flushleft}
        \footnotesize
        \hspace{2.1cm}\textsc{\aqua}\hspace{3.9cm}\textsc{\logiqa}\hspace{3.0cm}\textsc{\truthqa}
    \end{flushleft}
    \centering
    \begin{subfigure}[b]{0.32\textwidth}
        \centering
        \includegraphics[width=\linewidth]{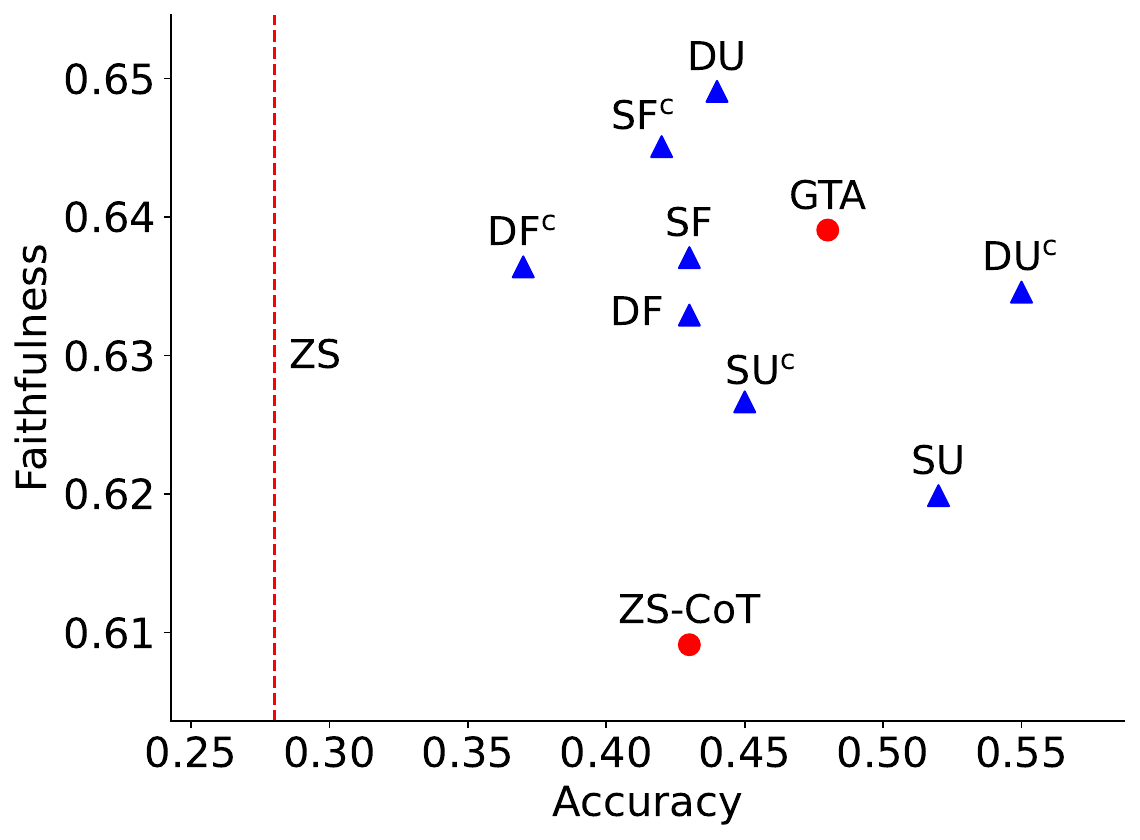}
    \end{subfigure}
    \hfill
    \begin{subfigure}[b]{0.32\textwidth}
        \centering
        \includegraphics[width=\linewidth]{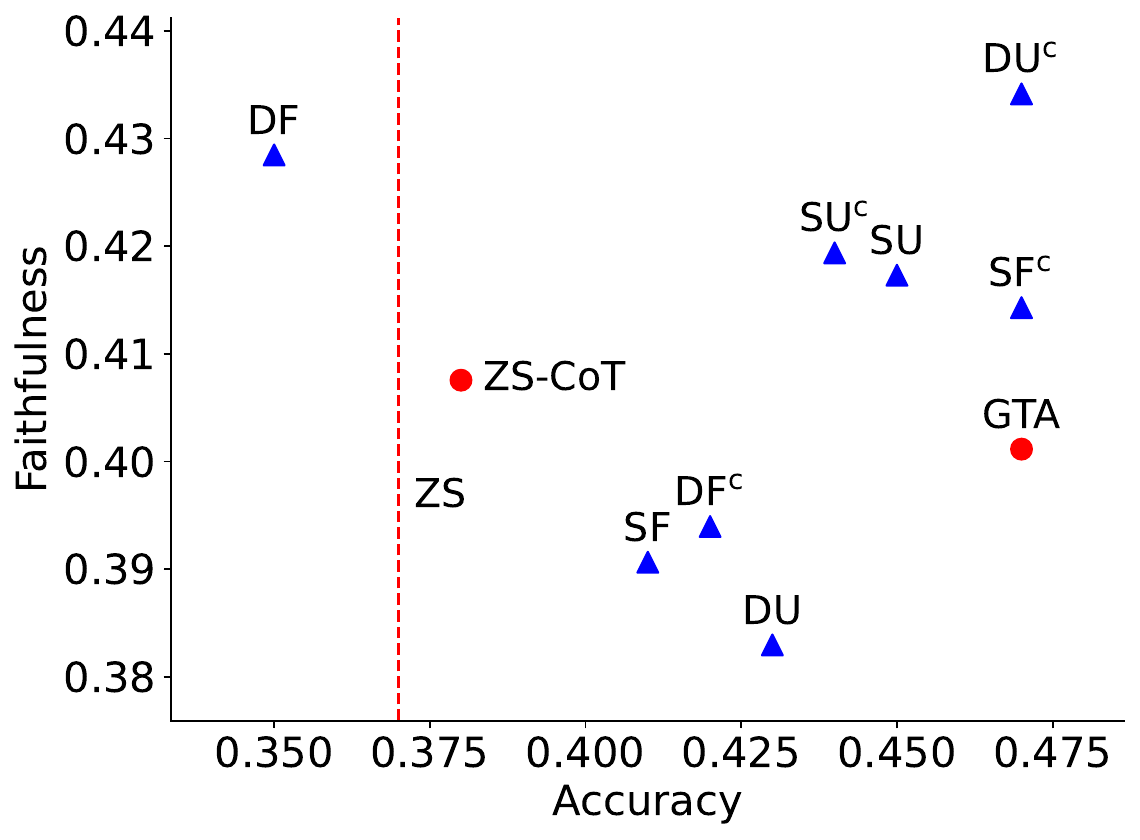}
    \end{subfigure}
    \hfill
    \begin{subfigure}[b]{0.32\textwidth}
        \centering
        \includegraphics[width=\linewidth]{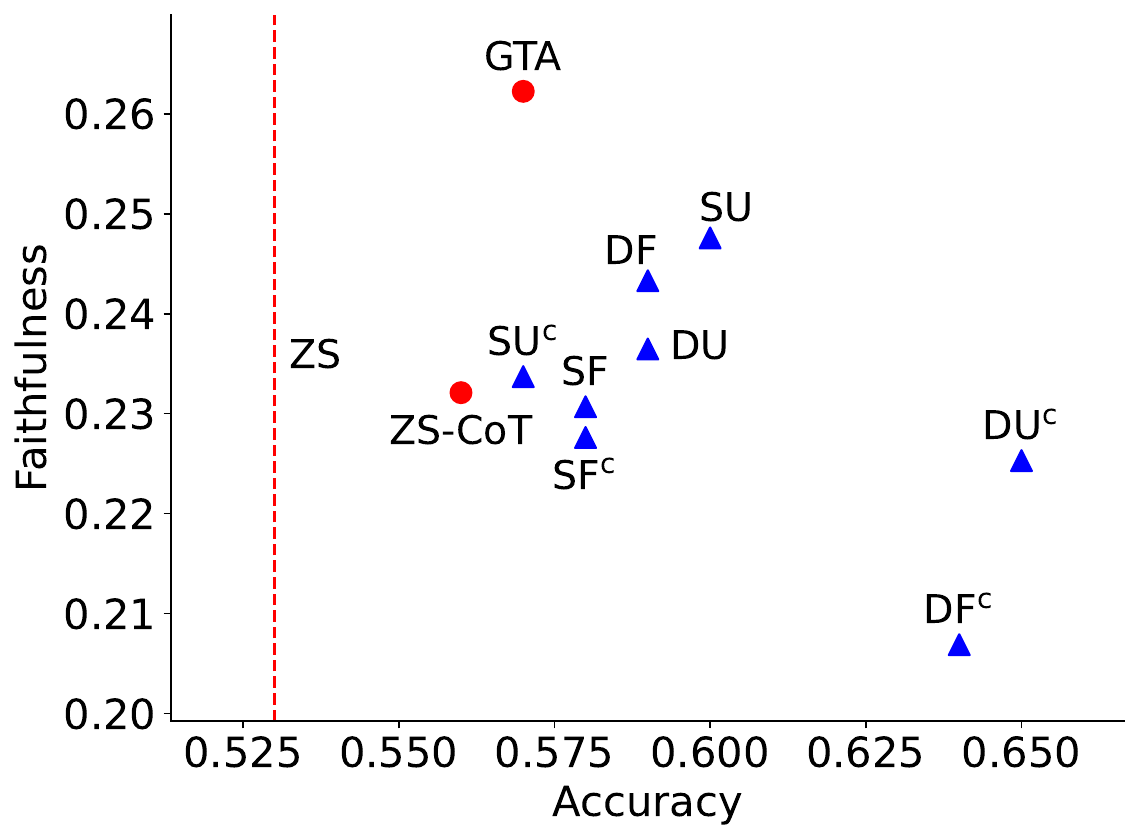}
    \end{subfigure}
    \vspace{-0.1in}
    \caption{\footnotesize{\looseness=-1 Faithfulness vs Accuracy relationship of CoT reasoning generated by \llama using different baseline (in red) and \textbf{ICL} strategies (in blue). Results show that none of the baseline or sampling strategy consistently achieve high accuracy and faithfulness.}
    }
    \vspace{-0.1in}
    \label{fig:icl_llama_faith_acc}
\end{figure}

\xhdr{More accurate LLMs are less faithful} On average, across three datasets, we find that \gptfour achieves significantly higher accuracy on all three datasets as compared to \gptthree and \llama (see Figs.~\ref{fig:icl_gpt_4_faith_acc},\ref{fig:icl_gpt_faith_acc},\ref{fig:icl_llama_faith_acc}), but it exhibits poor faithfulness performance. For instance, in \truthqa, we find that \gptfour provides correct answers to questions without using CoT reasoning (\ie accuracy difference between non-CoT and CoT prompting is zero), resulting in low faithfulness by definition. Also, larger LLMs like \gptfour are increasingly optimized for dialogue and generating conversational responses where RLHF rewards coherence to a human evaluator, which may conflict with generating faithful CoT reasoning.

\xhdr{In-context learning (ICL) improves faithfulness, albeit with a trade-off in accuracy} On all datasets and models, we observe that in-context learning improves faithfulness compared to zero shot baseline for almost all sampling strategies as shown in Figs.~\ref{fig:icl_gpt_4_faith_acc},\ref{fig:icl_gpt_faith_acc},\ref{fig:icl_llama_faith_acc}. Using faithful samples in-context particularly enhances faithfulness, as evidenced by a rise in faithfulness compared to the uniform counterpart, \ie faithfulness of DF > DU and SF > SU. One exception is \llama on \truthqa dataset. We suspect this is due to \truthqa being a dataset of human falsehoods relies less on reasoning to arrive at an answer. While ICL improves faithfulness, this often comes with a drop in accuracy as shown in Figs.~\ref{fig:icl_gpt_4_faith_acc},\ref{fig:icl_gpt_faith_acc},\ref{fig:icl_llama_faith_acc}.

\xhdr{Certain sampling strategies provide better trade-offs} Using top-K faithful samples (DF), on average, improves the faithfulness of the CoT reasoning but takes a hit on the accuracy, whereas the stochastic uniform sampling (SU) obtains better accuracy without improving faithfulness. Stochastic faithful sampling (SF) provides a middle ground. Moreover, we find better accuracy-faithfulness trade-offs when we perform ICL prompting using only the CoT reasoning from correctly predicted question-answer pairs by the LLM.

In summary, our results show that we cannot elicit faithful CoT reasoning from LLMs by simply using examples from different ICL strategies during inference without sacrificing accuracy.

\subsubsection{Fine-tuning Analysis} Here, we aim to investigate the possible benefits of fine-tuning techniques to elicit more faithful CoT reasoning from LLMs. We fine-tune \llama and \gptthree models\footnote{Due to OpenAI API errors at the time of experimentation~\cite{API_ERROR}, we were unable to access or evaluate fine-tuned versions of GPT-4.} using different baselines (Sec.~\ref{sec:setup}) and sampling techniques (Sec.~\ref{sec:finetune}). 

\begin{figure}[h]
    \begin{flushleft}
        \footnotesize
        \hspace{2.1cm}\textsc{\aqua}\hspace{3.9cm}\textsc{\logiqa}\hspace{3.0cm}\textsc{\truthqa}
    \end{flushleft}
    \centering
    \begin{subfigure}[b]{0.32\textwidth}
        \centering
        \includegraphics[width=\linewidth]{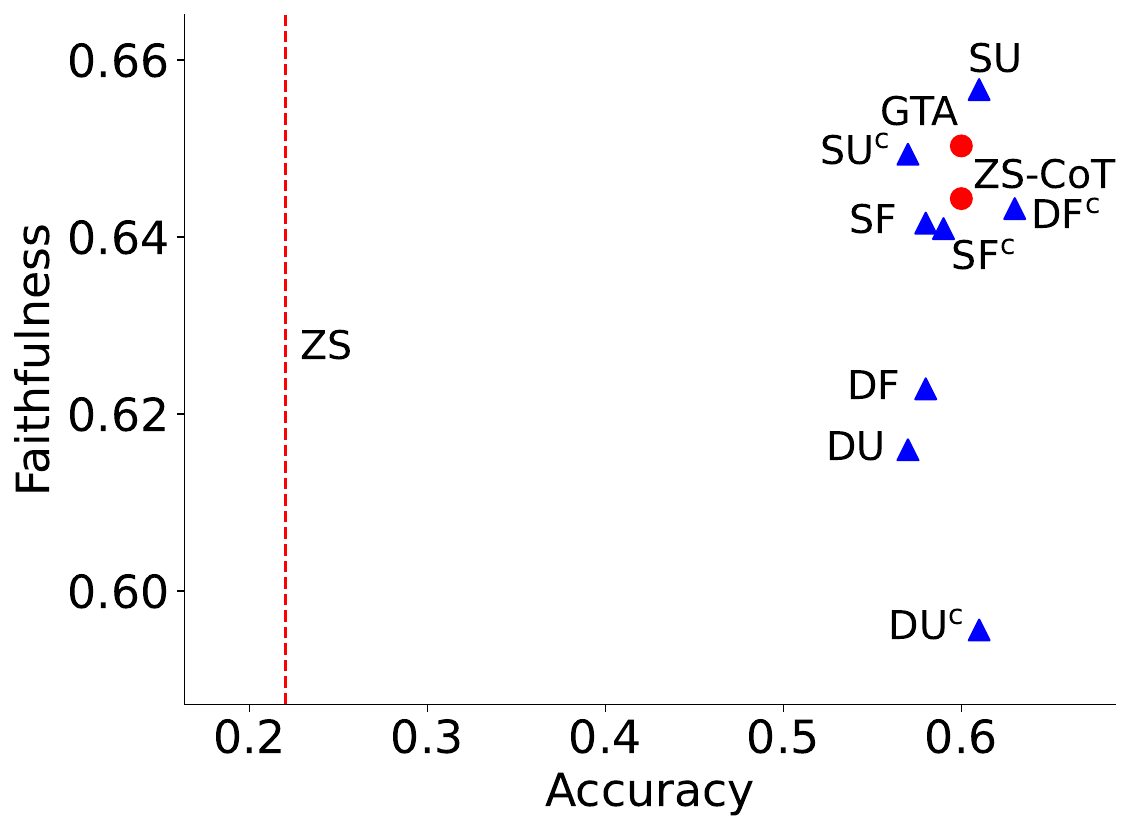}
        \label{fig:subfig1}
    \end{subfigure}
    \hfill
    \begin{subfigure}[b]{0.32\textwidth}
        \centering
        \includegraphics[width=\linewidth]{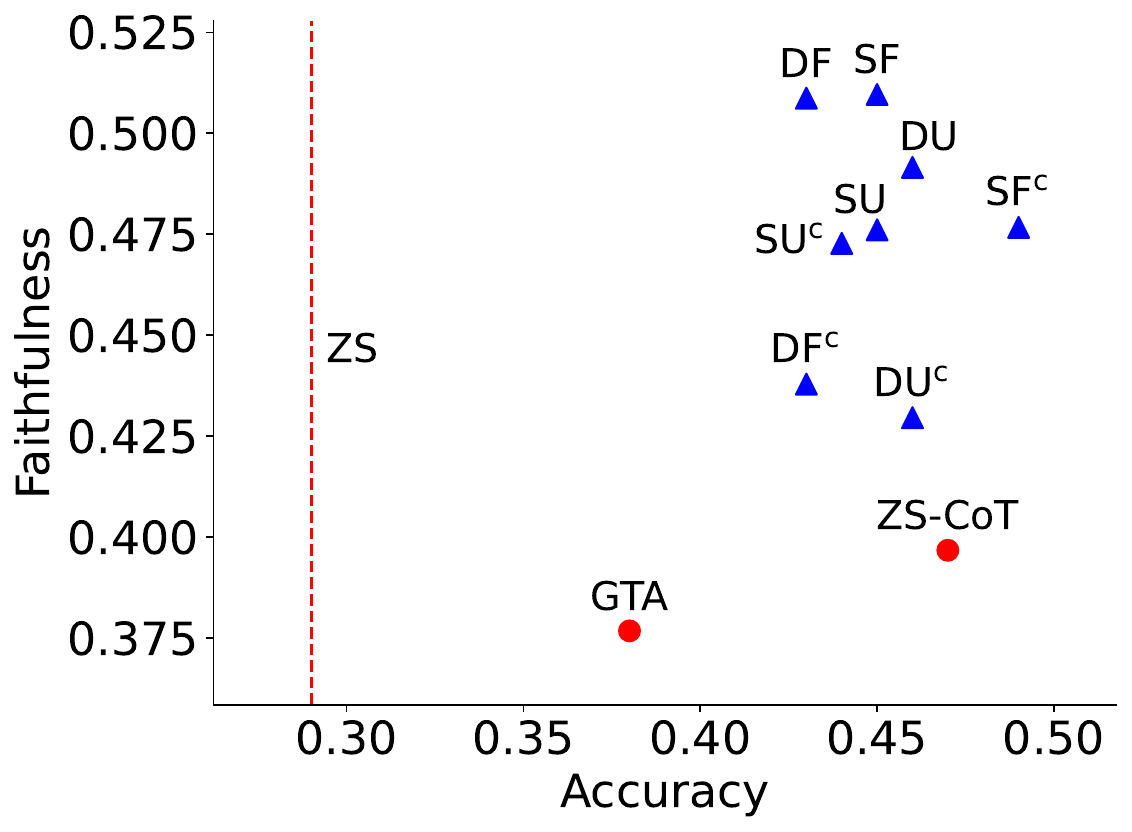}
        \label{fig:subfig2}
    \end{subfigure}
    \hfill
    \begin{subfigure}[b]{0.32\textwidth}
        \centering
        \includegraphics[width=\linewidth]{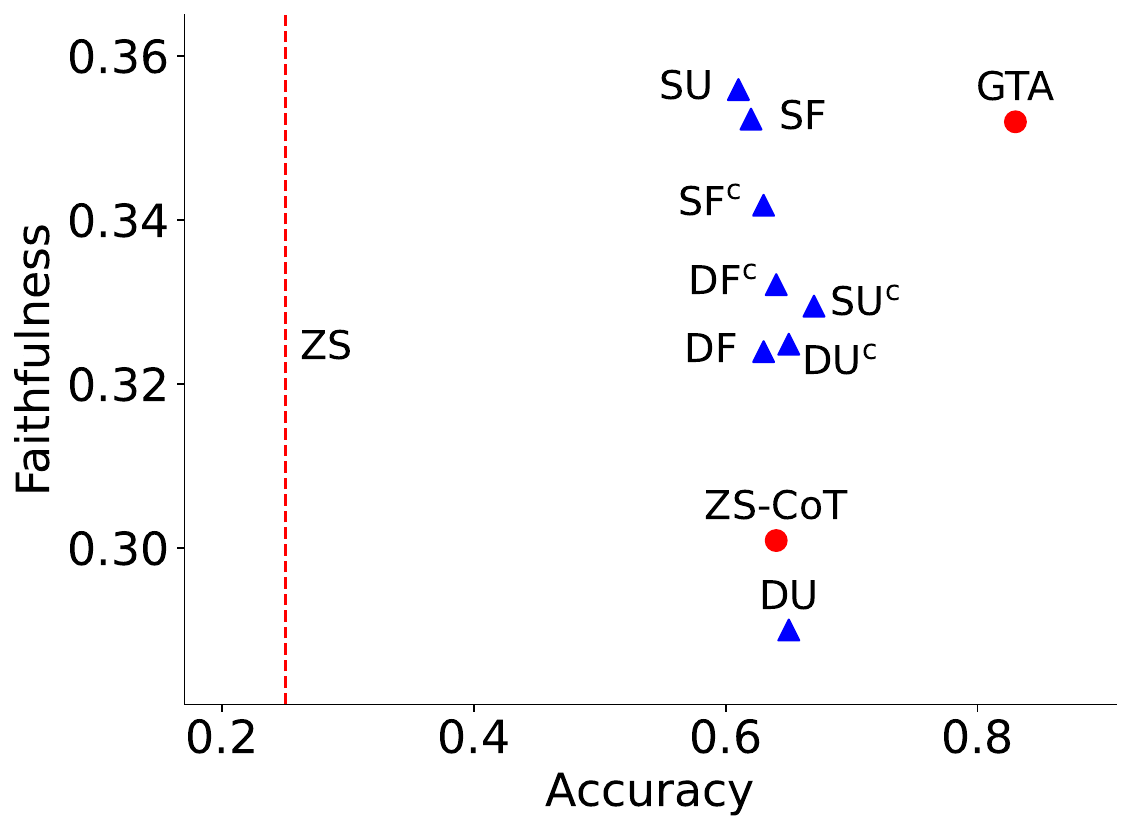}
        \label{fig:subfig3}
    \end{subfigure}
    \vspace{-0.15in}
    \caption{\footnotesize{\looseness=-1 Faithfulness vs Accuracy relationship of CoT reasoning generated by \textbf{fine-tuned} \gptthree using different baselines (in red) and sampling strategies (in blue). Results show that while the baseline \textit{GTA} achieves good accuracy-faithfulness trade-off (top-right corner) for \aqua and \truthqa dataset, it achieves the worst trade-off (bottom-left corner) for \logiqa dataset.}
    }
    \vspace{-0.1in}
    \label{fig:ft_gpt_faith_acc}
\end{figure}

\begin{figure}[h]
    \begin{flushleft}
        \footnotesize
        \hspace{2.1cm}\textsc{\aqua}\hspace{3.9cm}\textsc{\logiqa}\hspace{3.0cm}\textsc{\truthqa}
    \end{flushleft}
    \centering
    \begin{subfigure}[b]{0.32\textwidth}
        \centering
        \includegraphics[width=\linewidth]{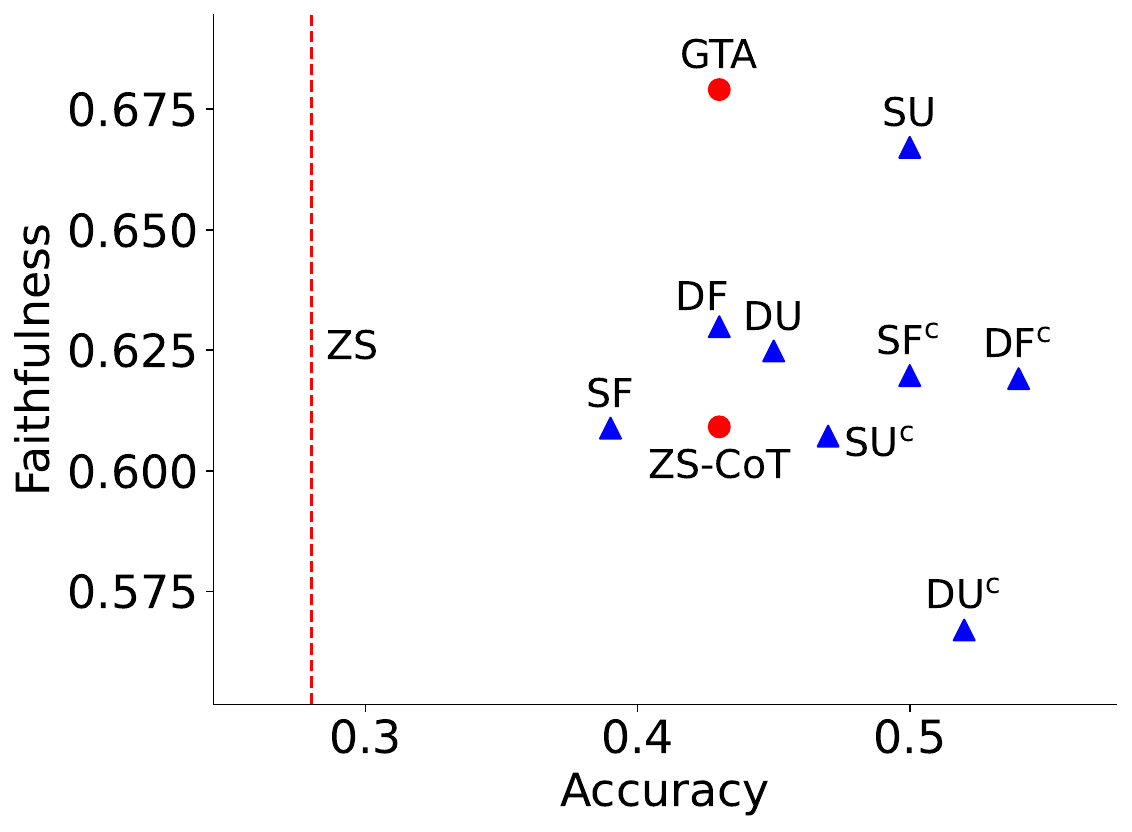}
    \end{subfigure}
    \hfill
    \begin{subfigure}[b]{0.32\textwidth}
        \centering
        \includegraphics[width=\linewidth]{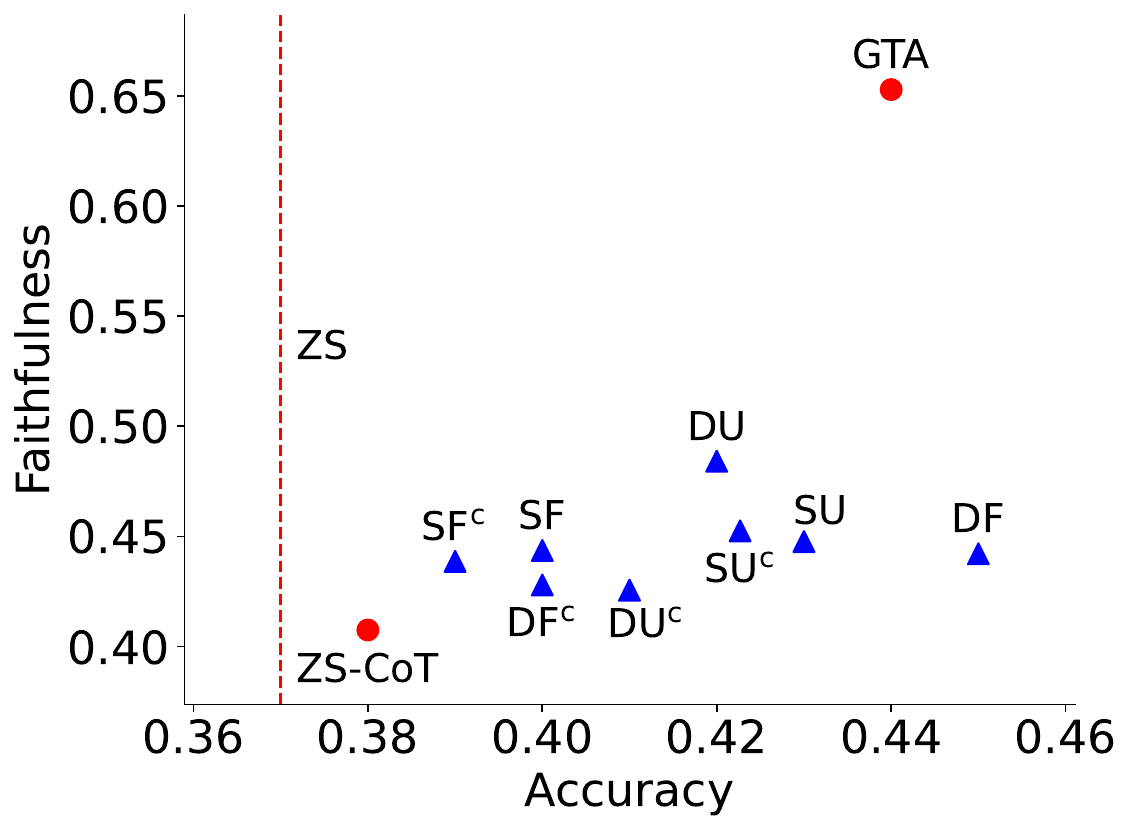}
    \end{subfigure}
    \hfill
    \begin{subfigure}[b]{0.32\textwidth}
        \centering
        \includegraphics[width=\linewidth]{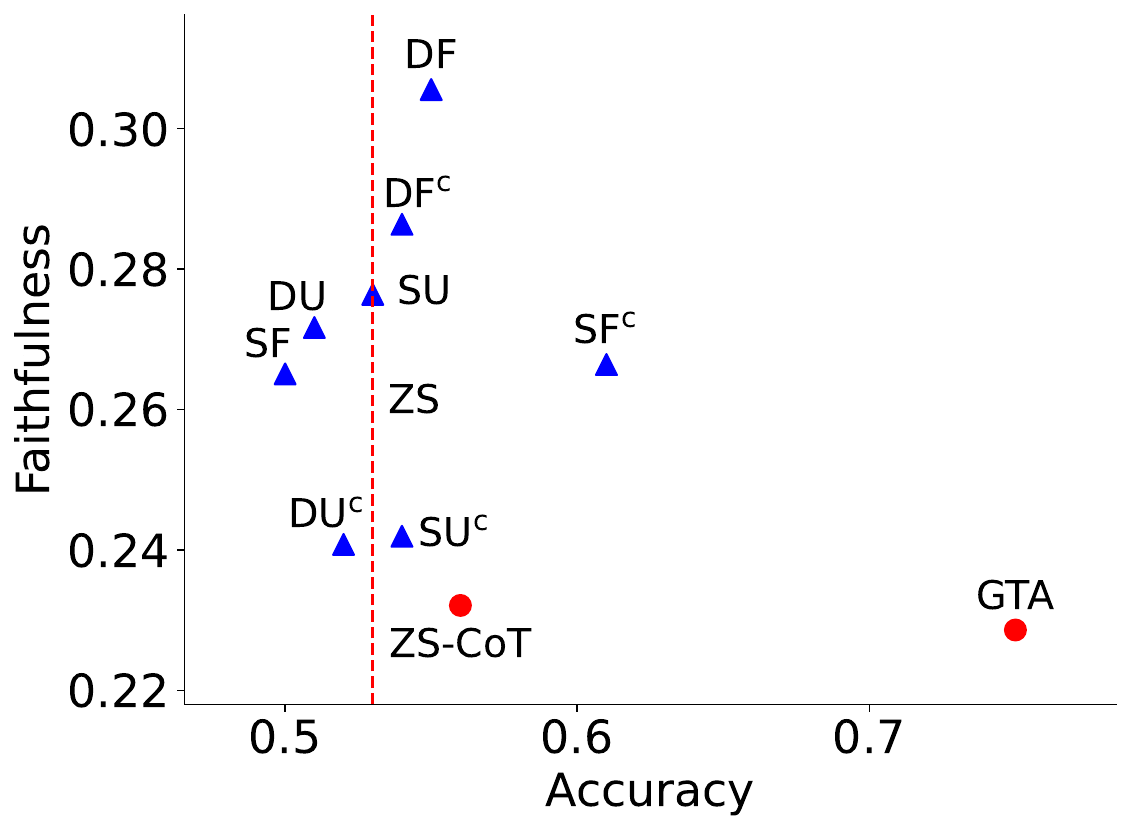}
    \end{subfigure}
    \vspace{-0.05in}
    \caption{\footnotesize{Faithfulness vs Accuracy relationship of CoT reasoning generated by \textbf{fine-tuned} \llama using different baselines (in red) and sampling strategies (in blue).
    On average, across all datasets, we find that none of the baseline or sampling strategies achieve high faithfulness.
    }
    }
    \vspace{-0.15in}
    \label{app:llama3-ft}
\end{figure}

\xhdr{Fine-tuned LLMs show contrasting faithfulness performance} Our results in Figs.~\ref{fig:ft_gpt_faith_acc} and \ref{app:llama3-ft} for \aqua and \logiqa show that while some sampling strategies lead to improvement in faithfulness of CoT reasoning for fine-tuned \gptthree, they obtain lower faithfulness than \textit{GTA} baseline for fine-tuned \llama. In addition, we observe that the baseline \textit{GTA} achieves a good accuracy-faithfulness trade-off for the \logiqa dataset (Fig.~\ref{app:llama3-ft}), it does not follow the same trend for fine-tuned \gptthree (Fig.~\ref{fig:ft_gpt_faith_acc}). Further, our fine-tuning results on \truthqa show that while we can force an LLM to generate faithful CoT reasoning via fine-tuning (verified by an increase in their faithfulness performance), it significantly impacts the accuracy of the model ($\sim$20\% drop in accuracy) (see \truthqa; Fig.~\ref{app:llama3-ft}).

\xhdr{Fine-tuning using most faithful explanations achieve better accuracy-faithfulness trade-offs} For the fine-tuned \gptthree on \logiqa dataset, we find that sampling strategies like $\textsc{DF}$ and $\textsc{SF}$ achieve higher faithfulness as compared to the baselines (in red), highlighting that selecting examples with faithful explanations for fine-tuning can help in generating faithful CoT reasoning from the fine-tuned LLMs. Notably, we observe a better accuracy-faithfulness trade-offs when fine-tuning using only the correctly predicted question-answer pairs with CoT reasoning (see Fig.~\ref{fig:ft_gpt_faith_acc}; $\text{DF}^\text{c}$ in \aqua and $\text{SF}^\text{c}$ in \logiqa).

\subsubsection{Activation Editing Analysis} 
\begin{figure}[h]
    \vspace{-0.05in}
    \begin{flushleft}
        \footnotesize
        \hspace{2.9cm}\textsc{\aqua}\hspace{5.8cm}\textsc{\truthqa}
    \end{flushleft}
    \vspace{-0.05in}
    \centering
    \includegraphics[width=\textwidth]{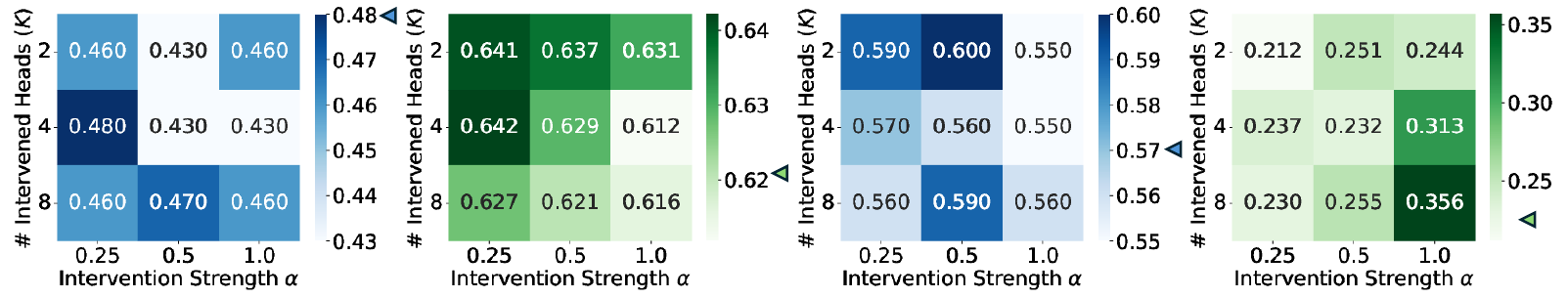}
    \begin{flushleft}
        \footnotesize
        \vspace{-0.08in}
        \hspace{1cm}\textsc{Accuracy}\hspace{1.8cm}\textsc{Faithfulness}\hspace{1.5cm}\textsc{Accuracy}\hspace{1.8cm}\textsc{Faithfulness}
    \end{flushleft}    
    \vspace{-0.1in}
    \caption{\footnotesize{\looseness=-1 Accuracy and Faithfulness of LLM reasoning for different intervention configurations ($\alpha, K$). The difference between the accuracy and faithfulness performance of \llama and highlights that none of the intervention configuration leads to improvement of both accuracy and faithfulness across both datasets compared to ZS-CoT performance (\trianbox1{cblue} and \trianbox1{cgreen} markers). Refer to Appendix Fig.~\ref{app:intervention} for \logiqa dataset.}
    }
    \vspace{-0.1in}
    \label{fig:intervention}
\end{figure}

Through activation editing, we aim to understand the effect of intervening on a model to amplify faithful behavior. The intervention equation described in \ref{eq:interventionequation} has a hyperparameter $\alpha$ indicating the intervention strength $\alpha$. Furthermore, we intervene only on the top-$K$ faithful heads (as identified in Fig.~\ref{fig:attentionheads}) in order to be minimally invasive. The results in Fig.~\ref{fig:intervention} show faithfulness and accuracy on \truthqa and \aqua upon intervening on different number of attention heads of \llama, \ie $K = \{2, 4, 8\}$, and intervention strengths, \ie $\alpha = \{0.25, 0.50, 1.0\}$.

\looseness=-1\xhdr{Intervening on attention heads leads to a drop in accuracy with a marginal gain in faithfulness} The results in Fig.~\ref{fig:intervention} show that intervening on the most faithful attention heads of \llama doesn't yield a significant boost in the faithfulness of its CoT reasoning. Interestingly, as compared to the ZS-CoT performance of \llama (\aqua: \{Accuracy: 0.49; Faithfulness: 0.627\} and \truthqa: \{Accuracy: 0.57; Faithfulness: 0.232\}), we find no significant improvement in both accuracy (Fig.~\ref{fig:intervention}; columns (a),(c)) and faithfulness (Fig.~\ref{fig:intervention}; columns (b),(d)). Moreover, the identified faithful attention heads, optimal value of intervention strength ($\alpha$), and optimal number of intervened heads ($K$) are not consistent across different datasets, highlighting the lack of generalization of activation editing strategies to various datasets. In addition, our analysis demonstrates contrasting behaviors in \llama, where activation editing works for improving truthfulness but shows mixed results for faithfulness, underscoring the challenge of eliciting faithful CoT reasoning from LLMs. Finally, our results also highlight the dichotomy between accuracy and faithfulness, where the values of $\{\alpha, K\}$ for the most faithful attention head (dark green in Fig.~\ref{fig:intervention}) are not always equivalent to the most accurate one (dark blue in Fig.~\ref{fig:intervention}). 

\section{Conclusion}
\label{sec:conclusion}
In this study, we investigated the challenge of eliciting faithfulness chain-of-thought reasoning in Large Language Models (LLMs). We explored three widely used techniques: activation editing, fine-tuning, and in-context learning (ICL) in our empirical analysis. Our results indicate that while these methods provided marginal improvements, none were sufficient to consistently enhance the CoT faithfulness across diverse datasets and LLMs. Our findings highlight the critical need for novel methodologies and a deeper understanding of LLMs' internal reasoning processes to generate more faithful CoT explanations.

\newpage
\bibliographystyle{abbrvnat}
\bibliography{ref}

\pagebreak

\appendix
\pagebreak
\section*{Appendix}
\label{sec:appendix}
\section{Broader Impact}
\label{sec:broaderimpact}
Our work focuses on exploring whether we can improve the faithfulness of the CoT reasoning generated by state-of-the-art LLMs. This has significant positive implications for societal benefit. For instance, if CoT reasoning output by LLMs faithfully captures the underlying model behavior, decision-makers and relevant stakeholders can leverage this to determine if, when, and how much to rely on the recommendations provided by LLMs. Therefore, our exploration itself is very valuable and has a substantial positive societal impact. Our analyses and findings indicate that existing techniques commonly used to steer behavior in LLMs are not effective in enhancing the faithfulness of LLM-generated CoT reasoning. While this finding is not particularly positive, we believe it is a step in the right direction, informing us of the complexity of the problem and underscoring the need for fundamentally different frameworks to address it. As far as we understand, our work does not have any potential negative societal impacts, as it is mainly an exploration to improve the faithfulness of LLM-generated CoT reasoning. 

\section{Related Work}

\paragraph{Chain-of-Thought Reasoning} Large Language Models (LLMs) produce Chain-of-Thought (CoT) reasoning~\cite{wei2022chain,agarwal2024faithfulness} to help provide end users with a peak into the reasoning process leading up to their response. While the CoT reasoning generated by these models is often appealing to human end users~\cite{wei2022chain,krishna2024post}, prior research has argued that LLM-generated CoT reasoning does not \emph{faithfully} capture the underlying behavior of these models and that this is a critical challenge particularly in applications involving high-stakes decision making~\cite{agarwal2024faithfulness}. 
For instance, as discussed in~\citet{agarwal2024faithfulness},  a doctor would benefit from seeing an explanation that faithfully captures why an LLM is recommending a particular diagnosis for a patient, as opposed to seeing some plausible explanation that could lead to the diagnosis at hand. In the former case, the doctor can actually use this faithful explanation to determine if and how much to rely on the model's recommendation. 

\paragraph{Evaluating the Faithfulness of CoT Reasoning} Despite the criticality of the faithfulness of LLM-generated CoT reasoning, there is very little work on analyzing and measuring this aspect of LLMs.~\citet{turpin2023language} were the first to demonstrate that CoT explanations may not faithfully capture the behavior of the underlying models. They showed that these explanations can be heavily influenced by biasing model inputs \eg by reordering multiple-choice options in a few-shot prompt to always make the answer “(A)”—which these models systematically fail to mention in their explanations.~\citet{lanham2023measuring} extended the above work and proposed novel metrics to measure the faithfulness of an LLM-generated CoT explanation. For instance, they propose an \emph{early answering} metric, which considers a generated CoT to be faithful if truncating that CoT causes the model to change its final response. Similarly, if \emph{adding mistakes} in a generated CoT causes the model to change its final response, then the original CoT can be considered faithful. Analogously, they proposed other metrics to measure faithfulness based on \emph{paraphrasing} the beginning portions of the original CoT as well as replacing the CoT with \emph{filler} tokens (\eg ellipses). Using these metrics, they demonstrated that the CoT reasoning produced by state-of-the-art LLMs does not faithfully capture the behavior of the underlying models. 

\paragraph{Enhancing the Quality of CoT Reasoning} While there are some prior works that tackled the problem of improving the quality of CoT reasoning~\cite{lyu2023faithful}, their focus was on improving its quality vis-a-vis human knowledge or understanding. For example,~\citet{lyu2023faithful} focused on generating a reasoning chain that could then be put through a deterministic math solver, and the resulting answer from this solver was compared to the answer produced by the LLM. The reasoning chain was considered to be faithful if the answers of the solver and the LLM matched. Note that this approach does not account for ensuring that the internal computations or the underlying behavior of the LLM was captured in the reasoning chain, which is the focus of our work. 

In summary, our work makes one of the initial attempts at exploring the promise of various popular paradigms, namely, activation editing, fine-tuning, and in-context learning, to improve the faithfulness of the CoT reasoning generated by LLMs. 

\section{Experiments}
\label{app:experiments}
Here, we provide additional results of our experiments in tabular format and perform significance testing of all our empirical analysis.
\begin{figure}[t]
    \begin{flushleft}
        \footnotesize
        \hspace{2.1cm}\textsc{\aqua}\hspace{3.9cm}\textsc{\logiqa}\hspace{3.0cm}\textsc{\truthqa}
    \end{flushleft}
    \centering
    \begin{subfigure}[b]{0.32\textwidth}
        \centering
        \includegraphics[width=\linewidth]{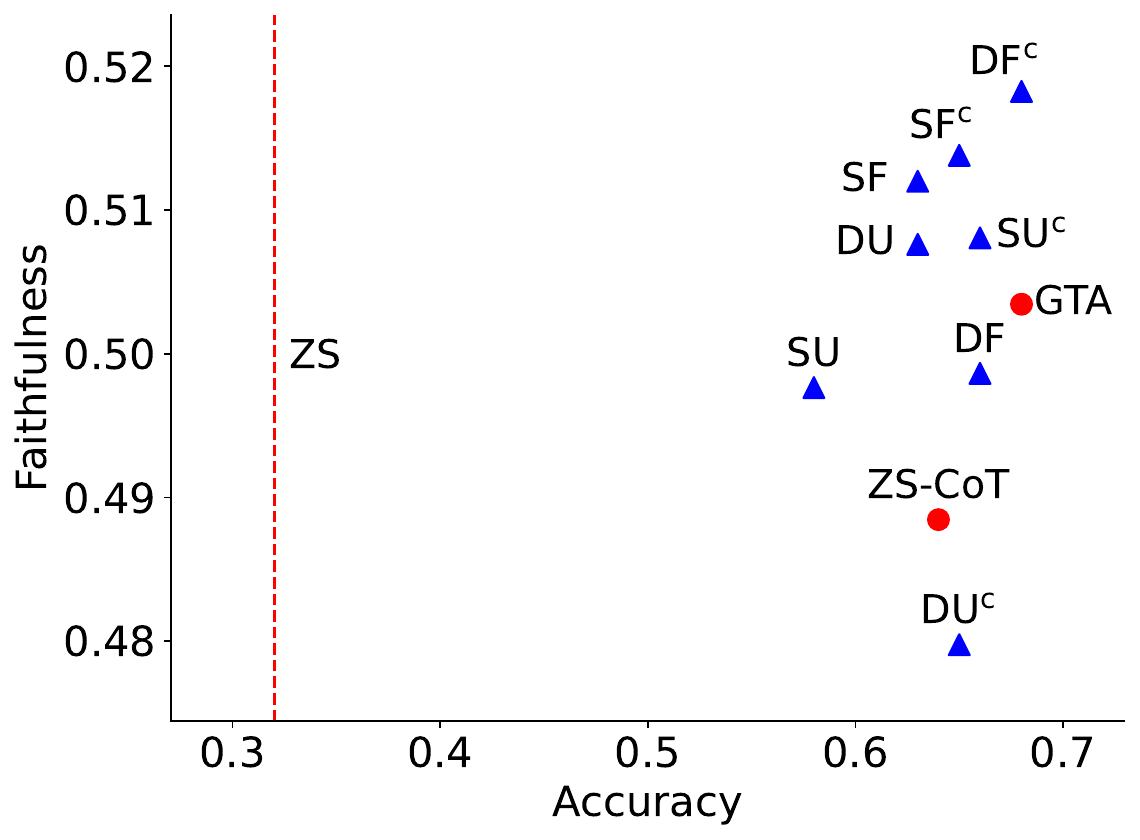}
        \label{fig:subfig1}
    \end{subfigure}
    \hfill
    \begin{subfigure}[b]{0.32\textwidth}
        \centering
        \includegraphics[width=\linewidth]{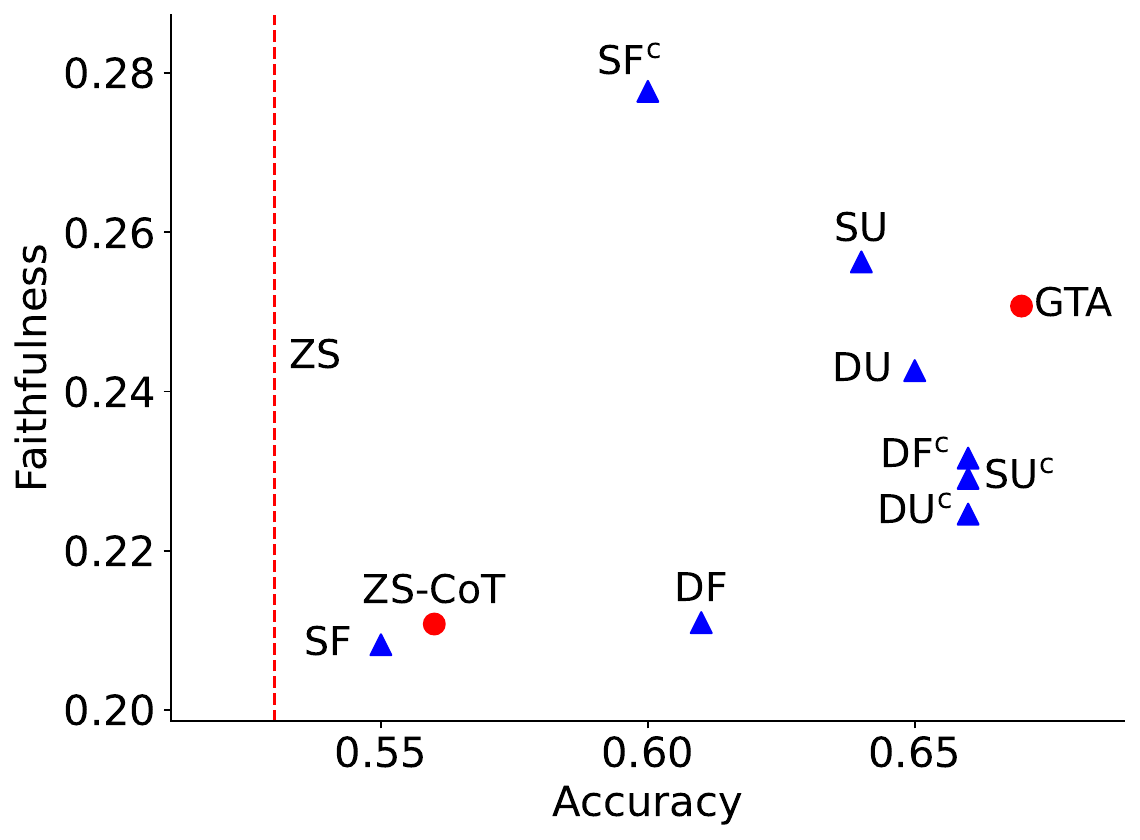}
        \label{fig:subfig2}
    \end{subfigure}
    \hfill
    \begin{subfigure}[b]{0.32\textwidth}
        \centering
        \includegraphics[width=\linewidth]{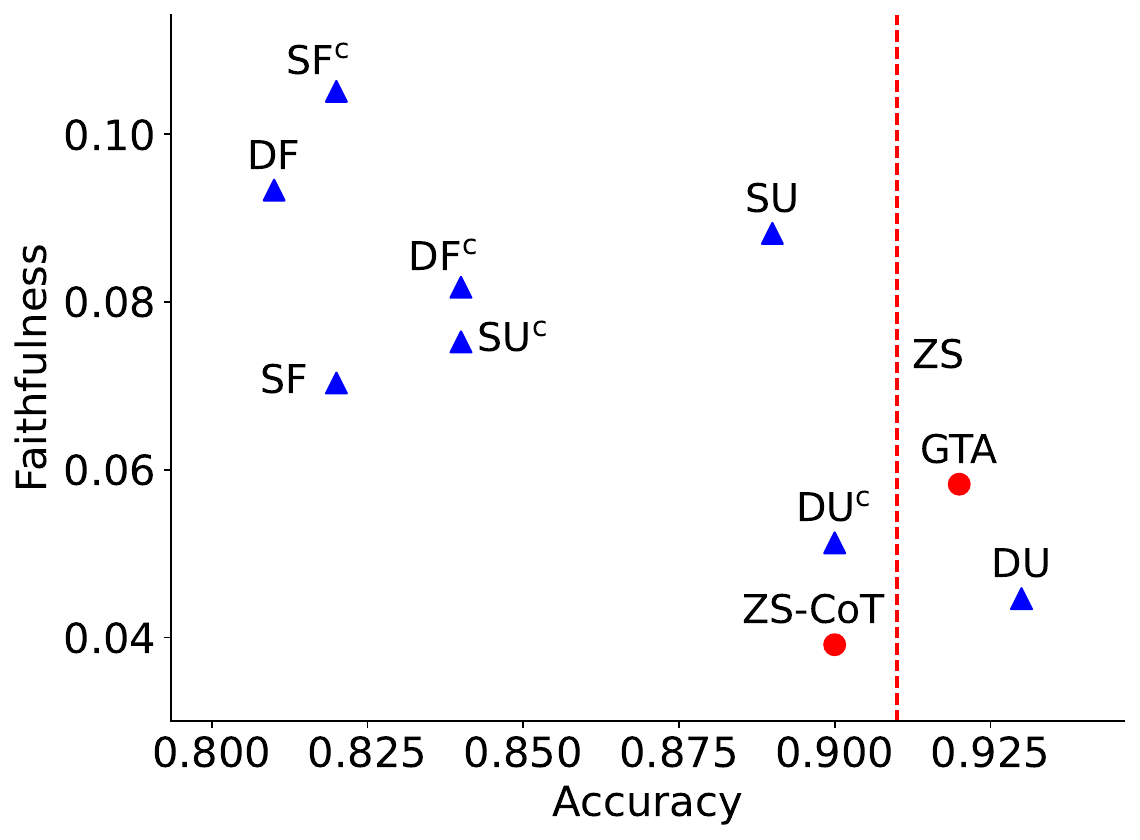}
        \label{fig:subfig3}
    \end{subfigure}
    \vspace{-0.15in}
    \caption{\looseness=-1 Faithfulness vs Accuracy relationship of CoT reasoning generated by \gptfour using different baseline (in red) and \textbf{ICL} strategies (in blue). Results show that stochastic faithful sampling strategies, on average across three datasets, achieves higher faithfulness in CoT reasoning.
    }
    \label{fig:icl_gpt_4_faith_acc}
\end{figure}

\begin{figure}[t]
    \centering
    \includegraphics[width=\textwidth]{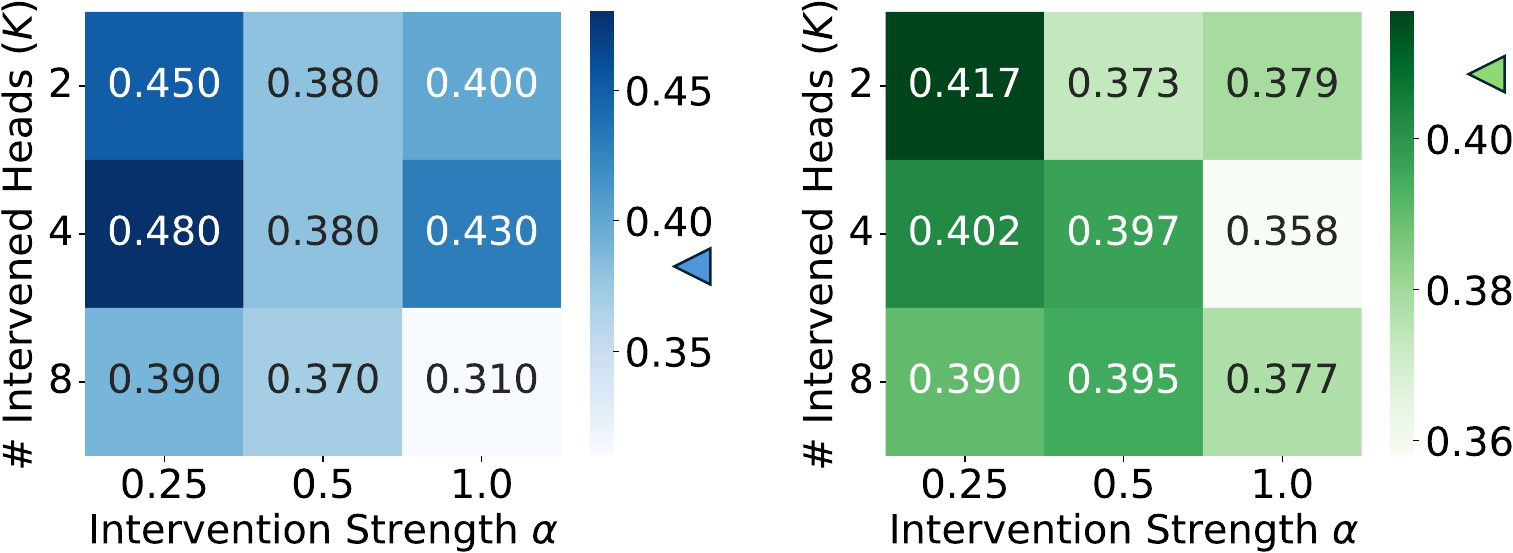}
   \begin{flushleft}
        \footnotesize
        \vspace{-0.05in}
        \hspace{2.1cm}\textsc{Accuracy}\hspace{6cm}\textsc{Faithfulness}
    \end{flushleft}   
    \caption{Accuracy and faithfulness of LLM reasoning for different intervention configurations ($\alpha, K$) for \logiqa dataset. Activation editing shows different the trade-off between the accuracy and faithfulness performance of \llama and some configuration leads to an increase in accuracy as compared to the zero-shot CoT performance (\trianbox1{cblue} and \trianbox1{cgreen} markers) but doesn't improve faithfulness significantly.
    }
    \label{app:intervention}
\end{figure}


\begin{table}[htbp]
    \centering
    \caption{GPT-3.5-Turbo Faithfulness for Different Fine-tuning Approaches}
    \label{tab:ft_gpt_faith_acc}
    \begin{tabular}{l c c c c c c}
        \toprule
        \textbf{Approach} & \multicolumn{2}{c}{\textbf{AQuA}} & \multicolumn{2}{c}{\textbf{LogiQA}} & \multicolumn{2}{c}{\textbf{TruthfulQA}}\\
         & Accuracy & Faithfulness & Accuracy & Faithfulness & Accuracy & Faithfulness \\
        \midrule
        ZS-CoT & 0.60 $\pm$ 0.05 & 0.64 $\pm$ 0.02 & 0.47 $\pm$ 0.05 & 0.40 $\pm$ 0.03 & 0.64 $\pm$ 0.05 & 0.30 $\pm$ 0.02\\
        GTA & 0.60 $\pm$ 0.05 & 0.65 $\pm$ 0.02 & 0.38 $\pm$ 0.05 & 0.38 $\pm$ 0.03 & \textbf{0.83} $\pm$ 0.04 & 0.35 $\pm$ 0.03\\ \midrule
        DU & 0.57 $\pm$ 0.05 & 0.62 $\pm$ 0.02 & 0.46 $\pm$ 0.05 & 0.49 $\pm$ 0.03 & 0.65 $\pm$ 0.05 & 0.29 $\pm$ 0.03\\
        DU$^\mathrm{c}$ & 0.61 $\pm$ 0.05 & 0.60 $\pm$ 0.03 & 0.46 $\pm$ 0.05 & 0.43 $\pm$ 0.03 & 0.65 $\pm$ 0.05 & 0.32 $\pm$ 0.03\\
        DF & 0.58 $\pm$ 0.05 & 0.62 $\pm$ 0.02 & 0.43 $\pm$ 0.05 & 0.51 $\pm$ 0.03 & 0.63 $\pm$ 0.05 & 0.32 $\pm$ 0.03\\
        DF$^\mathrm{c}$ & \textbf{0.63} $\pm$ 0.05 & 0.64 $\pm$ 0.02 & 0.43 $\pm$ 0.05 & 0.44 $\pm$ 0.03 & 0.64 $\pm$ 0.05 & 0.33 $\pm$ 0.03\\
        SU & 0.61 $\pm$ 0.05 & \textbf{0.66} $\pm$ 0.02 & 0.45 $\pm$ 0.05 & 0.48 $\pm$ 0.03 & 0.61 $\pm$ 0.05 & \textbf{0.36} $\pm$ 0.03\\
        SU$^\mathrm{c}$ & 0.57 $\pm$ 0.05 & 0.65 $\pm$ 0.02 & 0.44 $\pm$ 0.05 & 0.47 $\pm$ 0.03 & 0.67 $\pm$ 0.05 & 0.33 $\pm$ 0.02\\
        SF & 0.58 $\pm$ 0.05 & 0.64 $\pm$ 0.02 & 0.45 $\pm$ 0.05 & \textbf{0.51} $\pm$ 0.02 & 0.62 $\pm$ 0.05 & 0.35 $\pm$ 0.03\\
        SF$^\mathrm{c}$ & 0.59 $\pm$ 0.05 & 0.64 $\pm$ 0.02 & \textbf{0.49} $\pm$ 0.05 & 0.48 $\pm$ 0.03 & 0.63 $\pm$ 0.05 & 0.34 $\pm$ 0.02\\
        \bottomrule
    \end{tabular}
\end{table}

\begin{table}[htbp]
    \centering
    \caption{Llama-3-8B-Instruct Faithfulness for Different Fine-tuning Approaches}
    \label{tab:ft_llama_faith_acc}
    \begin{tabular}{l c c c c c c}
        \toprule
        \textbf{Approach} & \multicolumn{2}{c}{\textbf{AQuA}} & \multicolumn{2}{c}{\textbf{LogiQA}} & \multicolumn{2}{c}{\textbf{TruthfulQA}}\\
         & Accuracy & Faithfulness & Accuracy & Faithfulness & Accuracy & Faithfulness \\
        \midrule
        ZS-CoT & 0.43 $\pm$ 0.05 & 0.61 $\pm$ 0.02 & 0.38 $\pm$ 0.05 & 0.41 $\pm$ 0.03 & 0.56 $\pm$ 0.05 & 0.23 $\pm$ 0.03\\
        GTA & 0.43 $\pm$ 0.05 & \textbf{0.68} $\pm$ 0.01 & 0.44 $\pm$ 0.05 & \textbf{0.65} $\pm$ 0.01 & \textbf{0.75} $\pm$ 0.04 & 0.23 $\pm$ 0.03\\ \midrule
        DU & 0.45 $\pm$ 0.05 & 0.62 $\pm$ 0.02 & 0.42 $\pm$ 0.05 & 0.48 $\pm$ 0.02 & 0.51 $\pm$ 0.05 & 0.27 $\pm$ 0.03\\
        DU$^\mathrm{c}$ & 0.52 $\pm$ 0.05 & 0.57 $\pm$ 0.02 & 0.41 $\pm$ 0.05 & 0.43 $\pm$ 0.03 & 0.52 $\pm$ 0.05 & 0.24 $\pm$ 0.03\\
        DF & 0.43 $\pm$ 0.05 & 0.63 $\pm$ 0.02 & \textbf{0.45} $\pm$ 0.05 & 0.44 $\pm$ 0.02 & 0.55 $\pm$ 0.05 & \textbf{0.31} $\pm$ 0.03\\
        DF$^\mathrm{c}$ & \textbf{0.54} $\pm$ 0.05 & 0.62 $\pm$ 0.02 & 0.40 $\pm$ 0.05 & 0.43 $\pm$ 0.03 & 0.54 $\pm$ 0.05 & 0.29 $\pm$ 0.03\\
        SU & 0.50 $\pm$ 0.05 & 0.67 $\pm$ 0.01 & 0.43 $\pm$ 0.05 & 0.45 $\pm$ 0.03 & 0.53 $\pm$ 0.05 & 0.28 $\pm$ 0.03\\
        SU$^\mathrm{c}$ & 0.47 $\pm$ 0.05 & 0.61 $\pm$ 0.02 & 0.42 $\pm$ 0.05 & 0.45 $\pm$ 0.03 & 0.54 $\pm$ 0.05 & 0.24 $\pm$ 0.03\\
        SF & 0.39 $\pm$ 0.05 & 0.61 $\pm$ 0.02 & 0.40 $\pm$ 0.05 & 0.44 $\pm$ 0.03 & 0.50 $\pm$ 0.05 & 0.27 $\pm$ 0.03\\
        SF$^\mathrm{c}$ & 0.50 $\pm$ 0.05 & 0.62 $\pm$ 0.02 & 0.39 $\pm$ 0.05 & 0.44 $\pm$ 0.03 & 0.61 $\pm$ 0.05 & 0.27 $\pm$ 0.03\\

        \bottomrule
    \end{tabular}
\end{table}

\begin{table}[htbp]
    \centering
    \caption{GPT-4 Faithfulness for Different In-Context Learning Approaches}
    \label{tab:icl_gpt4_faith_acc}
    \begin{tabular}{l c c c c c c}
        \toprule
        \textbf{Approach} & \multicolumn{2}{c}{\textbf{AQuA}} & \multicolumn{2}{c}{\textbf{LogiQA}} & \multicolumn{2}{c}{\textbf{TruthfulQA}}\\
         & Accuracy & Faithfulness & Accuracy & Faithfulness & Accuracy & Faithfulness \\
        \midrule
        ZS-CoT & 0.64 $\pm$ 0.05 & 0.49 $\pm$ 0.03 & 0.56 $\pm$ 0.05 & 0.21 $\pm$ 0.03 & 0.90 $\pm$ 0.03 & 0.04 $\pm$ 0.01\\
        GTA & \textbf{0.68} $\pm$ 0.05 & 0.50 $\pm$ 0.03 & \textbf{0.67} $\pm$ 0.05 & 0.25 $\pm$ 0.03 & 0.92 $\pm$ 0.03 & 0.06 $\pm$ 0.02\\ \midrule
        DU & 0.63 $\pm$ 0.05 & 0.51 $\pm$ 0.03 & 0.65 $\pm$ 0.05 & 0.24 $\pm$ 0.03 & \textbf{0.93} $\pm$ 0.03 & 0.04 $\pm$ 0.01\\
        DU$^\mathrm{c}$ & 0.65 $\pm$ 0.05 & 0.48 $\pm$ 0.03 & 0.66 $\pm$ 0.05 & 0.22 $\pm$ 0.02 & 0.90 $\pm$ 0.03 & 0.05 $\pm$ 0.01\\
        DF & 0.66 $\pm$ 0.05 & 0.50 $\pm$ 0.03 & 0.61 $\pm$ 0.05 & 0.21 $\pm$ 0.02 & 0.81 $\pm$ 0.04 & 0.09 $\pm$ 0.02\\
        DF$^\mathrm{c}$ & \textbf{0.68} $\pm$ 0.05 & \textbf{0.52} $\pm$ 0.03 & 0.66 $\pm$ 0.05 & 0.23 $\pm$ 0.03 & 0.84 $\pm$ 0.04 & 0.08 $\pm$ 0.02\\
        SU & 0.58 $\pm$ 0.05 & 0.50 $\pm$ 0.03 & 0.64 $\pm$ 0.05 & 0.26 $\pm$ 0.03 & 0.89 $\pm$ 0.03 & 0.09 $\pm$ 0.02\\
        SU$^\mathrm{c}$ & 0.66 $\pm$ 0.05 & 0.51 $\pm$ 0.03 & 0.66 $\pm$ 0.05 & 0.23 $\pm$ 0.03 & 0.84 $\pm$ 0.04 & 0.08 $\pm$ 0.02\\
        SF & 0.63 $\pm$ 0.05 & 0.51 $\pm$ 0.03 & 0.55 $\pm$ 0.05 & 0.21 $\pm$ 0.02 & 0.82 $\pm$ 0.04 & 0.07 $\pm$ 0.02\\
        SF$^\mathrm{c}$ & 0.65 $\pm$ 0.05 & 0.51 $\pm$ 0.03 & 0.60 $\pm$ 0.05 & \textbf{0.28} $\pm$ 0.03 & 0.82 $\pm$ 0.04 & \textbf{0.11} $\pm$ 0.02\\
        \bottomrule
    \end{tabular}
\end{table}

\begin{table}[htbp]
    \centering
    \caption{GPT-3.5-Turbo Faithfulness for Different In-Context Learning Approaches}
    \label{tab:icl_gpt_faith_acc}
    \begin{tabular}{l c c c c c c}
        \toprule
        \textbf{Approach} & \multicolumn{2}{c}{\textbf{AQuA}} & \multicolumn{2}{c}{\textbf{LogiQA}} & \multicolumn{2}{c}{\textbf{TruthfulQA}}\\
         & Accuracy & Faithfulness & Accuracy & Faithfulness & Accuracy & Faithfulness \\
        \midrule
        ZS-CoT & 0.60 $\pm$ 0.05 & 0.64 $\pm$ 0.02 & 0.47 $\pm$ 0.05 & 0.40 $\pm$ 0.03 & 0.64 $\pm$ 0.05 & 0.30 $\pm$ 0.02\\
        GTA & 0.56 $\pm$ 0.05 & 0.64 $\pm$ 0.02 & 0.48 $\pm$ 0.05 & 0.44 $\pm$ 0.03 & 0.69 $\pm$ 0.05 & 0.33 $\pm$ 0.02\\
        DU & 0.55 $\pm$ 0.05 & 0.67 $\pm$ 0.02 & 0.48 $\pm$ 0.05 & 0.40 $\pm$ 0.03 & 0.65 $\pm$ 0.05 & 0.32 $\pm$ 0.02\\ \midrule
        DU$^\mathrm{c}$ & \textbf{0.64} $\pm$ 0.05 & 0.64 $\pm$ 0.02 & 0.40 $\pm$ 0.05 & 0.44 $\pm$ 0.03 & 0.75 $\pm$ 0.04 & 0.30 $\pm$ 0.03\\
        DF & 0.57 $\pm$ 0.05 & 0.67 $\pm$ 0.02 & \textbf{0.54} $\pm$ 0.05 & \textbf{0.47} $\pm$ 0.03 & 0.74 $\pm$ 0.04 & 0.30 $\pm$ 0.02\\
        DF$^\mathrm{c}$ & 0.62 $\pm$ 0.05 & 0.65 $\pm$ 0.02 & 0.44 $\pm$ 0.05 & 0.45 $\pm$ 0.03 & 0.73 $\pm$ 0.04 & \textbf{0.34} $\pm$ 0.03\\
        SU & 0.59 $\pm$ 0.05 & 0.66 $\pm$ 0.02 & 0.43 $\pm$ 0.05 & 0.46 $\pm$ 0.03 & \textbf{0.78} $\pm$ 0.04 & 0.30 $\pm$ 0.03\\
        SU$^\mathrm{c}$ & 0.61 $\pm$ 0.05 & 0.65 $\pm$ 0.02 & 0.44 $\pm$ 0.05 & 0.45 $\pm$ 0.03 & 0.73 $\pm$ 0.04 & 0.32 $\pm$ 0.02\\
        SF & 0.59 $\pm$ 0.05 & \textbf{0.67} $\pm$ 0.02 & 0.48 $\pm$ 0.05 & 0.46 $\pm$ 0.03 & 0.70 $\pm$ 0.05 & 0.30 $\pm$ 0.02\\
        SF$^\mathrm{c}$ & 0.62 $\pm$ 0.05 & 0.66 $\pm$ 0.02 & 0.39 $\pm$ 0.05 & 0.44 $\pm$ 0.03 & 0.70 $\pm$ 0.05 & 0.31 $\pm$ 0.03\\
        \bottomrule
    \end{tabular}
\end{table}

\begin{table}[htbp]
    \centering
    \caption{Llama-3-8B-Instruct Faithfulness for Different In-Context Learning Approaches}
    \label{tab:icl_llama_faith_acc}
    \begin{tabular}{l c c c c c c}
        \toprule
        \textbf{Approach} & \multicolumn{2}{c}{\textbf{AQuA}} & \multicolumn{2}{c}{\textbf{LogiQA}} & \multicolumn{2}{c}{\textbf{TruthfulQA}}\\
         & Accuracy & Faithfulness & Accuracy & Faithfulness & Accuracy & Faithfulness \\
        \midrule
        ZS-CoT & 0.43 $\pm$ 0.05 & 0.61 $\pm$ 0.02 & 0.38 $\pm$ 0.05 & 0.41 $\pm$ 0.03 & 0.56 $\pm$ 0.05 & 0.23 $\pm$ 0.03\\
        GTA & 0.48 $\pm$ 0.05 & 0.64 $\pm$ 0.02 & \textbf{0.47} $\pm$ 0.05 & 0.40 $\pm$ 0.03 & 0.57 $\pm$ 0.05 & \textbf{0.26} $\pm$ 0.03\\ \midrule
        DU & 0.44 $\pm$ 0.05 & \textbf{0.65} $\pm$ 0.02 & 0.43 $\pm$ 0.05 & 0.38 $\pm$ 0.03 & 0.59 $\pm$ 0.05 & 0.24 $\pm$ 0.03\\
        DU$^\mathrm{c}$ & \textbf{0.55} $\pm$ 0.05 & 0.63 $\pm$ 0.02 & \textbf{0.47} $\pm$ 0.05 & \textbf{0.43} $\pm$ 0.03 & \textbf{0.65} $\pm$ 0.05 & 0.23 $\pm$ 0.03\\
        DF & 0.43 $\pm$ 0.05 & 0.63 $\pm$ 0.02 & 0.35 $\pm$ 0.05 & 0.43 $\pm$ 0.03 & 0.59 $\pm$ 0.05 & 0.24 $\pm$ 0.03\\
        DF$^\mathrm{c}$ & 0.37 $\pm$ 0.05 & 0.64 $\pm$ 0.02 & 0.42 $\pm$ 0.05 & 0.39 $\pm$ 0.03 & 0.64 $\pm$ 0.05 & 0.21 $\pm$ 0.03\\
        SU & 0.52 $\pm$ 0.05 & 0.62 $\pm$ 0.02 & 0.45 $\pm$ 0.05 & 0.42 $\pm$ 0.03 & 0.60 $\pm$ 0.05 & 0.25 $\pm$ 0.03\\
        SU$^\mathrm{c}$ & 0.45 $\pm$ 0.05 & 0.63 $\pm$ 0.02 & 0.44 $\pm$ 0.05 & 0.42 $\pm$ 0.03 & 0.57 $\pm$ 0.05 & 0.23 $\pm$ 0.03\\
        SF & 0.43 $\pm$ 0.05 & 0.64 $\pm$ 0.02 & 0.41 $\pm$ 0.05 & 0.39 $\pm$ 0.03 & 0.58 $\pm$ 0.05 & 0.23 $\pm$ 0.03\\
        SF$^\mathrm{c}$ & 0.42 $\pm$ 0.05 & 0.65 $\pm$ 0.02 & \textbf{0.47} $\pm$ 0.05 & 0.41 $\pm$ 0.03 & 0.58 $\pm$ 0.05 & 0.23 $\pm$ 0.03\\
        \bottomrule
    \end{tabular}
\end{table}

\begin{table}[htbp]
    \centering
    \caption{GPT-3.5-Turbo p-values of Faithfulness for Different Fine-tuning Approaches}
    \label{tab:ft_gpt_p_values}
    \begin{tabular}{l c c c c c c}
        \toprule
        \textbf{Comparing} & \multicolumn{2}{c}{\textbf{AQuA}} & \multicolumn{2}{c}{\textbf{LogiQA}} & \multicolumn{2}{c}{\textbf{TruthfulQA}} \\
         & ZS-CoT & GTA & ZS-CoT & GTA & ZS-CoT & GTA \\
        \midrule
        DU & 0.1946 & 0.2247 & 0.0000 & 0.0005 & 0.6353 & 0.1101\\
        DU$^\mathrm{c}$ & 0.0718 & 0.0974 & 0.2141 & 0.0597 & 0.3573 & 0.4600\\
        DF & 0.3640 & 0.2610 & 0.0000 & 0.0000 & 0.3607 & 0.4292\\
        DF$^\mathrm{c}$ & 0.9523 & 0.7473 & 0.0917 & 0.0201 & 0.2364 & 0.6090\\
        SU & 0.4740 & 0.7740 & 0.0014 & 0.0010 & 0.0473 & 0.9173\\
        SU$^\mathrm{c}$ & 0.8063 & 0.9671 & 0.0088 & 0.0028 & 0.2353 & 0.5102\\
        SF & 0.8789 & 0.6707 & 0.0000 & 0.0000 & 0.0579 & 0.9934\\
        SF$^\mathrm{c}$ & 0.8324 & 0.6255 & 0.0006 & 0.0001 & 0.1071 & 0.7794\\
        \bottomrule
    \end{tabular}
\end{table}

\begin{table}[htbp]
    \centering
    \caption{Llama-3-8B-Instruct p-values for Different Fine-tuning Approaches}
    \label{tab:ft_llama_p_values}
    \begin{tabular}{l c c c c c c}
        \toprule
        \textbf{Comparing} & \multicolumn{2}{c}{\textbf{AQuA}} & \multicolumn{2}{c}{\textbf{LogiQA}} & \multicolumn{2}{c}{\textbf{TruthfulQA}} \\
         & ZS-CoT & GTA & ZS-CoT & GTA & ZS-CoT & GTA \\
        \midrule
        DU & 0.4325 & 0.0062 & 0.0027 & 0.0000 & 0.1835 & 0.1687\\
        DU$^\mathrm{c}$ & 0.0845 & 0.0000 & 0.5589 & 0.0000 & 0.7541 & 0.7380\\
        DF & 0.3175 & 0.0103 & 0.1958 & 0.0000 & 0.0130 & 0.0194\\
        DF$^\mathrm{c}$ & 0.6068 & 0.0011 & 0.3946 & 0.0000 & 0.0580 & 0.0639\\
        SU & 0.0020 & 0.4670 & 0.1636 & 0.0000 & 0.1311 & 0.1476\\
        SU$^\mathrm{c}$ & 0.9323 & 0.0020 & 0.1537 & 0.0000 & 0.7327 & 0.6844\\
        SF & 0.9893 & 0.0003 & 0.2321 & 0.0000 & 0.2940 & 0.2954\\
        SF$^\mathrm{c}$ & 0.6049 & 0.0012 & 0.3319 & 0.0000 & 0.1527 & 0.2178\\
        \bottomrule
    \end{tabular}
\end{table}

\begin{table}[htbp]
    \centering
    \caption{GPT-4 p-values for Different In-Context Learning Approaches}
    \label{tab:icl_gpt4_p_values}
    \begin{tabular}{l c c c c c c}
        \toprule
        \textbf{Comparing} & \multicolumn{2}{c}{\textbf{AQuA}} & \multicolumn{2}{c}{\textbf{LogiQA}} & \multicolumn{2}{c}{\textbf{TruthfulQA}} \\
         & ZS-CoT & GTA & ZS-CoT & GTA & ZS-CoT & GTA \\
        \midrule
        DU & 0.3089 & 0.8058 & 0.1395 & 0.7462 & 0.6632 & 0.2648\\
        DU$^\mathrm{c}$ & 0.6307 & 0.1890 & 0.4525 & 0.3024 & 0.2392 & 0.5765\\
        DF & 0.5638 & 0.7929 & 0.9936 & 0.1062 & 0.0048 & 0.0489\\
        DF$^\mathrm{c}$ & 0.1322 & 0.4820 & 0.3382 & 0.4337 & 0.0250 & 0.2369\\
        SU & 0.6778 & 0.7624 & 0.0509 & 0.8104 & 0.0063 & 0.0572\\
        SU$^\mathrm{c}$ & 0.3145 & 0.7599 & 0.3932 & 0.3125 & 0.0297 & 0.2525\\
        SF & 0.2818 & 0.6367 & 0.9038 & 0.0491 & 0.0478 & 0.5067\\
        SF$^\mathrm{c}$ & 0.2677 & 0.5417 & 0.0037 & 0.2679 & 0.0011 & 0.0111\\
        \bottomrule
    \end{tabular}
\end{table}

\begin{table}[htbp]
    \centering
    \caption{GPT-3.5-Turbo p-values for Different In-Context Learning Approaches}
    \label{tab:icl_gpt_p_values}
    \begin{tabular}{l c c c c c c}
        \toprule
        \textbf{Comparing} & \multicolumn{2}{c}{\textbf{AQuA}} & \multicolumn{2}{c}{\textbf{LogiQA}} & \multicolumn{2}{c}{\textbf{TruthfulQA}} \\
         & ZS-CoT & GTA & ZS-CoT & GTA & ZS-CoT & GTA \\
        \midrule
        DU & 0.2748 & 0.1188 & 0.8037 & 0.1245 & 0.4544 & 0.7770\\
        DU$^\mathrm{c}$ & 0.8994 & 0.8539 & 0.1285 & 0.8728 & 0.9518 & 0.3670\\
        DF & 0.1845 & 0.0451 & 0.0144 & 0.3093 & 0.9840 & 0.3429\\
        DF$^\mathrm{c}$ & 0.8065 & 0.5908 & 0.0505 & 0.7696 & 0.2248 & 0.7428\\
        SU & 0.4238 & 0.2524 & 0.0186 & 0.5463 & 0.9364 & 0.3364\\
        SU$^\mathrm{c}$ & 0.8541 & 0.5486 & 0.0323 & 0.6991 & 0.5434 & 0.6946\\
        SF & 0.0992 & 0.0558 & 0.0093 & 0.3931 & 0.8899 & 0.4127\\
        SF$^\mathrm{c}$ & 0.2790 & 0.1526 & 0.1431 & 0.8505 & 0.6492 & 0.6452\\
        \bottomrule
    \end{tabular}
\end{table}

\begin{table}[htbp]
    \centering
    \caption{Llama-3-8B-Instruct p-values for Different In-Context Learning Approaches}
    \label{tab:icl_llama_p_values}
    \begin{tabular}{l c c c c c c}
        \toprule
        \textbf{Comparing} & \multicolumn{2}{c}{\textbf{AQuA}} & \multicolumn{2}{c}{\textbf{LogiQA}} & \multicolumn{2}{c}{\textbf{TruthfulQA}} \\
         & ZS-CoT & GTA & ZS-CoT & GTA & ZS-CoT & GTA \\
        \midrule
        DU & 0.0488 & 0.5230 & 0.3320 & 0.4825 & 0.8569 & 0.2464\\
        DU$^\mathrm{c}$ & 0.2151 & 0.8029 & 0.3341 & 0.2573 & 0.7859 & 0.1824\\
        DF & 0.2610 & 0.7089 & 0.4776 & 0.3101 & 0.6582 & 0.4255\\
        DF$^\mathrm{c}$ & 0.2190 & 0.8713 & 0.6078 & 0.8081 & 0.3891 & 0.0104\\
        SU & 0.6302 & 0.2352 & 0.7058 & 0.5657 & 0.5822 & 0.5463\\
        SU$^\mathrm{c}$ & 0.3399 & 0.4633 & 0.6561 & 0.5424 & 0.9518 & 0.2365\\
        SF & 0.2268 & 0.8976 & 0.4570 & 0.7079 & 0.9604 & 0.2342\\
        SF$^\mathrm{c}$ & 0.1095 & 0.7537 & 0.8032 & 0.6608 & 0.8679 & 0.1552\\
        \bottomrule
    \end{tabular}
\end{table}

\end{document}